\documentclass[11pt]{article}

\usepackage[final]{acl}

\usepackage{times}
\usepackage{latexsym}

\usepackage[T1]{fontenc}

\usepackage[utf8]{inputenc}

\usepackage{microtype}

\usepackage{inconsolata}

\usepackage[ruled,vlined]{algorithm2e}
\usepackage{graphicx}

\usepackage{subcaption}
\usepackage{amsmath}
\usepackage{cleveref}
\usepackage{amsfonts}
\usepackage{booktabs}
\usepackage{tabularx}
\usepackage{caption}
\usepackage{graphicx}
\usepackage{fontawesome5}
\usepackage{multirow}
\usepackage{caption}
\usepackage{enumitem}
\usepackage{array}
\usepackage{arydshln}
\usepackage[utf8]{inputenc}

\newcommand{\modelname}{VIGiA}
\newcommand{\modellongname}{\textbf{V}isual and \textbf{I}nstructional \textbf{G}u\textbf{i}ding \textbf{A}ssistant}
\newcommand{\datasetname}{InstructionVidDial}

\newcommand{\fireemoji}{Trn}
\newcommand{\iceemoji}{Frz}

\title{VIGiA: Instructional Video Guidance via Dialogue Reasoning and Retrieval}

\author{Diogo Glória-Silva, David Semedo, João Magalhães \\
  NOVA LINCS, NOVA School of Science and Technology, Portugal \\
  \texttt{dmgc.silva@campus.fct.unl.pt}\\
  \texttt{\{df.semedo, jmag\}@fct.unl.pt
 }
}
\usepackage{hyperref}
\hypersetup{
  pdfcreator = {},
  pdfproducer = {}
}

\begin{document}

\maketitle
\begin{abstract}
We introduce \modelname{}, a novel multimodal dialogue model designed to understand and reason over complex, multi-step instructional video action plans. Unlike prior work which focuses mainly on text-only guidance, or treats vision and language in isolation, \modelname{} supports grounded, plan-aware dialogue that requires reasoning over visual inputs, instructional plans, and interleaved user interactions. To this end, \modelname{} incorporates two key capabilities: (1) multimodal plan reasoning, enabling the model to align uni- and multimodal queries with the current task plan and respond accurately; and (2) plan-based retrieval, allowing it to retrieve relevant plan steps in either textual or visual representations. 
  Experiments were done on a novel dataset with rich Instructional Video Dialogues aligned with Cooking and DIY plans. 
  Our evaluation shows that \modelname{} outperforms existing state-of-the-art models on all tasks in a conversational plan guidance setting, reaching over 90\% accuracy on plan-aware VQA.
  ~\footnote{Model, code and dataset can be found at \url{https://github.com/dmgcsilva/vigia}.}
\end{abstract}

\begin{figure}[!t]
\centering
   \includegraphics[width=0.45\textwidth, trim={120 245 10 70}, clip]{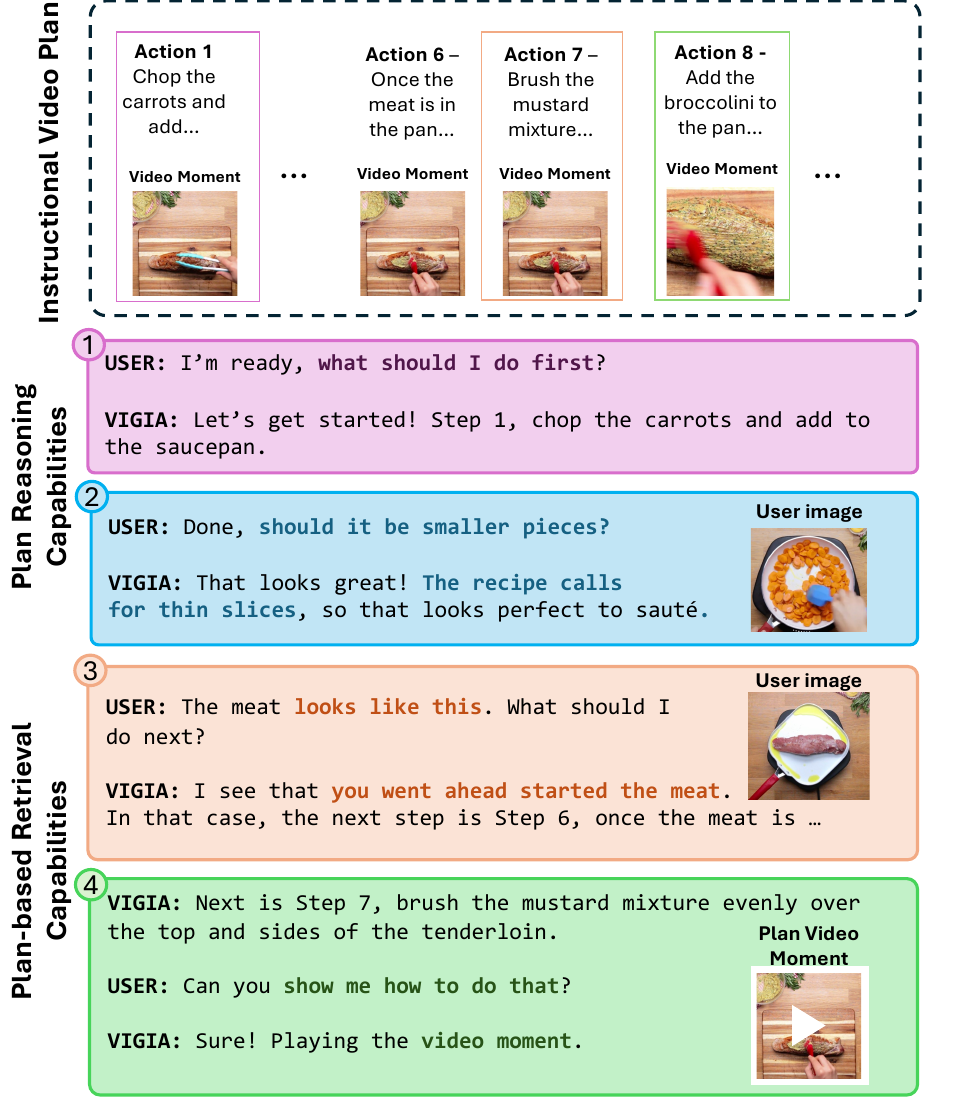}
   \caption{\modelname{} is grounded on complex multimodal instructional plans, delivering unified multimodal alignment over dialogue turns by providing text-based guidance (Goal 1), plan-aware visual question answering (Goal 2) aligning plan actions and visual context (Goal 3), and retrieving relevant video moments (Goal 4).}
   \label{fig:teaser}
   \vspace{-2mm}
\end{figure}

\section{Introduction}

Assisting in the execution of complex instructional plans (e.g. cooking and DIY projects), in a conversational manner, is a recent task~\cite{wizard_of_tasks} that requires Large Vision and Language Models (LVLMs) to understand and reason about logical sequences of actions.
As illustrated in Figure~\ref{fig:teaser}, this task presents a unique challenge due to the nature of the setting, where models need to be able to attend to various and complex instructions~\cite{wizard_of_tasks, fischer2024grillbot} arbitrarily interleaved throughout the dialogue, while remaining grounded in the instructional plan and tracking explicit and implicit plan progress.
Although some works address text-only plan guidance~\cite{actionabot, planllm, Zhu2025ActionaBotSM}, only a few consider multimodality~\cite{mmplanllm} and yet image-grounded reasoning tasks remain largely unexplored, a particularly challenging scenario, as it requires the model to ground its answer on the provided image, the instructional plan, the dialogue history, and its encoded knowledge.

\begin{figure*}[t!]
    \centering
    \begin{subfigure}[t]{0.33\textwidth}
        \centering
    \includegraphics[width=1\linewidth]{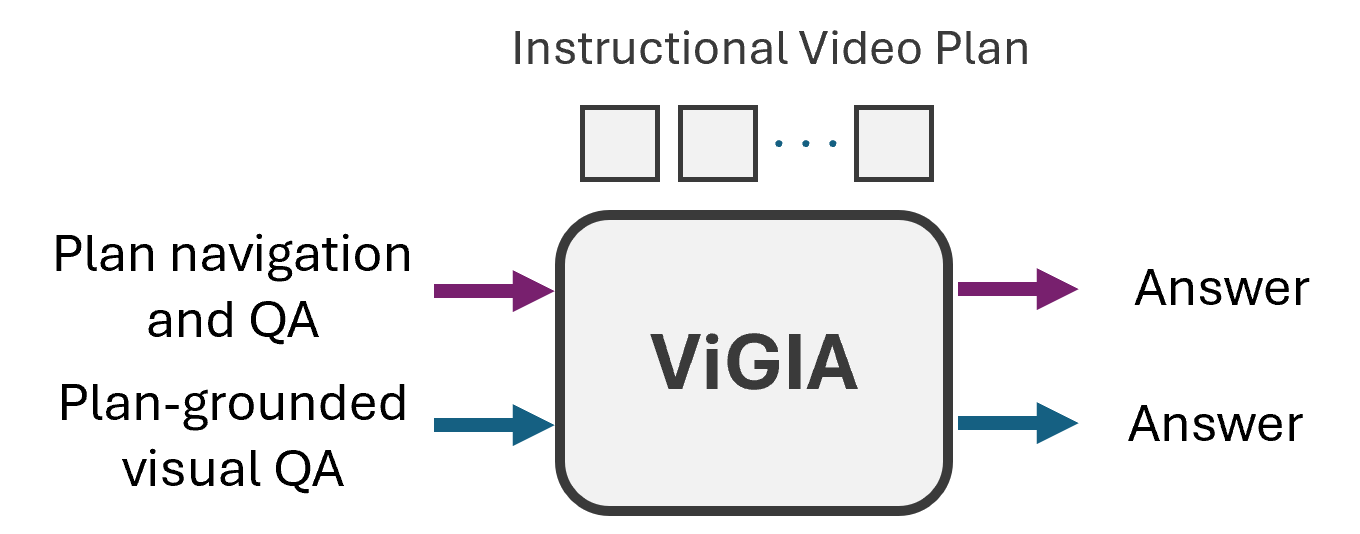}
        \caption{Plan reasoning capabilities.}
    \end{subfigure}%
    ~
    \begin{subfigure}[t]{0.65\textwidth}
        \centering
    \includegraphics[width=1\linewidth]{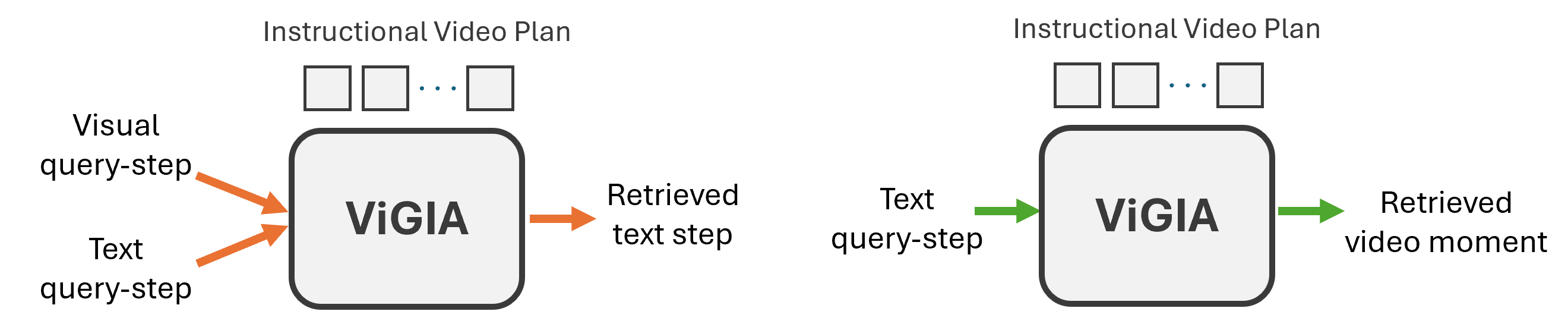}
        \caption{Plan-based retrieval capabilities.}
    \end{subfigure}
    \caption{\modelname{} is an LVLM model that processes instructional video plans to navigate through steps of the plan, perform QA, VQA and retrieval of arbitrarily random text steps or video moment steps.}
    \label{fig:capacities}
    \vspace{-2mm}
\end{figure*}

In this paper, we address the challenge in a comprehensive manner and propose \textbf{\modelname{}} (\modellongname{}), a complete conversational plan guidance model with strong multimodal support. To allow for a wide array of request types, \modelname{} has two key capability types: (1) \textbf{multimodal plan reasoning} capabilities and (2) \textbf{plan-based retrieval} capabilities. 
The first enables it to reason about uni- and multimodal requests, ground and align them on a multimodal plan, and answer accurately.
The second allows it to align multimodal input queries against the instructional plan and perform both in-context retrieval of a plan action text or retrieval of the video moment of that action.

To address both plan reasoning and retrieval capabilities, \modelname{}'s multi-goal architecture is purpose-built to support the multiple output and input modalities, and trained in a multi-objective learning paradigm using a novel dataset, \datasetname{}. This dataset incorporates all of these supported goals via semi-automatic dialogue generation and expands the scope by considering multiple plan domains. 
A multi-stage training approach progressively delivers \modelname{} capacity to follow complex plan-oriented multimodal instructions and support for multi-goal unified plan-guidance (Figure~\ref{fig:capacities}), generalizable to multiple domains.
Our main contributions are:
\begin{itemize}
    \item \textbf{\modelname{}}: A novel model that dialogues and reasons about multi-step instructional plans, while allowing for arbitrary retrieval of plan actions both in text and visual domain.
    \item \textbf{\datasetname{}}: We built a novel multimodal dialogue dataset that pushes plan-guidance into the complex and diverse domains of Cooking and DIY tasks.
\end{itemize}

Our evaluation compared \modelname{} to existing state-of-the-art models and results showed that it was the best model on plan-grounded reasoning QA and VQA tasks, while also delivering top performance across all plan-grounded retrieval tasks.

\section{Related Work}

LVLMs are models that are able to process not only text but also visual input (images and videos).
These models combine the strengths of language decoders with visual encoders, enabling multimodal input and, in some cases, multimodal output. LVLMs have benefited from the advances in pretrained Large Language Models~\cite{llama3, deepseekai2024deepseekv3technicalreport, gemma} as they can be paired with pretrained visual encoders such as CLIP~\cite{clip} or SigLIP~\cite{siglip} to form an LVLM connected via a connector module.
Several architectures have been proposed for the connector, ranging in complexity. While examples like the LLaVA family of models~\cite{llava, llava_rlhf, Liu2023-vr, llavar} use a linear or MLP module, others used more complex approaches, an example is the QFormer in BLIP models~\cite{blip_2, instructblip} or the Perceiver~\cite{perceiver} in Flamingo~\cite{flamingo}.

The need for modality equilibrium in LVLMs is key to their success, as performance in text-only settings needs to be maintained while also being able to attend to rich multimodal inputs. This led to increasingly more complex training approaches, with modern LVLMs training in a multistage paradigm~\cite{chen2024internvl, yang2024qwen2, Liu2023-vr, liu2024nvila, idefics2, idefics3}, allowing to progressively build up capabilities in a structured manner, while preventing catastrophic forgetting~\cite{mccloskey1989catastrophic} of previously learned tasks. The increased complexity of this has also led some works~\cite{idefics3, cambrian, apollo} to measure and verify the benefit of multistage training.

Instructional plan guidance is the multifaceted task of guiding users through complex manual instructional plans in a user-led conversational manner. Although early work has focused on a text-only setting~\cite{wizard_of_tasks, mo-etal-2023-roll, fischer2024grillbot, planllm}, these plans are inherently multimodal and often accompanied by instructional videos. The availability of large instructional video datasets~\cite{tasty_dataset, crosstask, coin} makes the application of LVLMs particularly suitable for this task. This was the case of MM-PlanLLM~\cite{mmplanllm}, however, it lacked the ability to answer image-grounded questions. 
While VQA has been studied and applied to several domains~\cite{Mathew2020DocVQAADA, Lobry2020RSVQAVQA, Ren2020CGMVQAANA, Tapaswi2015MovieQAUSA}, its application in instructional plan guidance remains unexplored.

\section{Definitions and Problem Formulation}

Let $P_k = \langle (a_{k,1}, vm_{k,1}), \cdots, (a_{k,m},vm_{k,m}) \rangle$ be the instructional plan  $k$, composed of actions  $a_{k,j}$ with the text action description, which is also illustrated by the video moment $vm_{k,m}$.
An instructional plan has an associated sequence of video moments $V_k = \langle vm_{k,1}, \cdots, vm_{k, m} \rangle$ illustrating the execution of the plan $P_k$.
Each video moment
$vm_{k,m} = \langle f_{k, \text{start}}, \dots, f_{k, \text{end}} \rangle \subseteq V_k$ range from a starting frame to an ending frame. Text actions and video moments are ordered in a logical sequence, where $a_{k,j}$ or $vm_{k,j}$ must be completed before $a_{k,j+1}$ or $vm_{k,j+1}$.

Let $D_k$ be a dialogue that is composed of multimodal interactions and associated with a specific instructional plan, and an instructional video that covers all actions. Formally, each dialogue instance is defined as $D_k = (P_k, T_k)$ where $T_k = \langle t_{k,1}, t_{k,2}, \cdots, t_{k,n} \rangle$ is the sequence of turns that constitutes the dialogue interaction for $d_k$. Each turn $t_{k,j}$ is defined as $t_{k,j} = (u_{k,j}, r_{k,j}, I_{k,j})$, where $u_{k,j}$ is the human input (question or request) during the $j$-th turn, $r_{k,j}$ is the system response, and $I_{k,j} \in \mathcal{I} \cup \{\texttt{None}\}$ is an optional image provided by the user during that turn.

This problem setting requires a  model that aligns the instructional plan with the dialogue context and accurately perform complex reasoning and plan retrieval operations. 

\section{Methodology}

\begin{figure*}[ht]
    \centering
    \includegraphics[width=1.0\textwidth, trim={7mm 0mm 7mm 0mm}, clip]{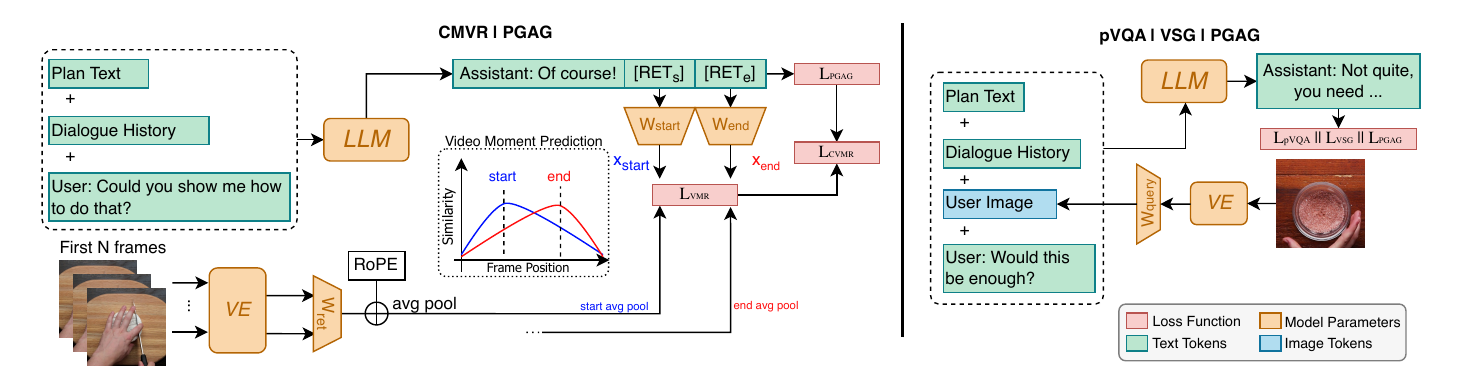}
     \caption{Global view of \modelname{}'s architecture. To handle multimodal inputs \modelname{} combines a visual encoder and an LLM using an MLP as a connector module. For conversational video moment retrieval, \modelname{} outputs a dedicated video moment start and end representation that can be used for start and end frame retrieval.}
     \label{fig:model_arch}
     \vspace{-2mm}
\end{figure*}

Addressing the problem outlined above, requires multimodal, context-aware, and plan-grounded reasoning, which will be integrated by design into \modelname{}'s architecture and training strategies, as detailed in the next sections. In addition, we will also detail the supporting \datasetname{} dataset.

\subsection{Multimodal Plan Guidance LVLMs}

Due to the required set of capabilities and input diversity, optimizing a single learning objective is infeasible. We address this by defining dedicated learning objectives for each type of dialogue capability.

\subsubsection{Plan-grounded reasoning capabilities}
The first set of dialogue capabilities support the core requests to navigate and interact with the instructional plan.

\paragraph{\textbf{Plan-Grounded Answer Generation.}} The task of textual plan-grounded answer generation encompasses every textual request that requires the generation of a response based on the dialogue and the instructional plan, such as plan navigation requests and questions. Here, a textual request $u_{k,j+1}$, a instructional plan $P_k$ and the previous conversation turns $T_{k,1:j} \in D_k$, the model needs to generate $r_{j+1} = \{ w_1, w_2, \cdots , w_t\}$ that accurately answers the request. In \modelname{}, the loss function for this task is the negative log-likelihood of the generated response tokens:
\begin{equation}
\small
    \mathcal{L}_{PGAG} = -\sum_{t=1} \log P( w_t | T_{k,1:j}, u_{k,j+1}, P_k)
\end{equation}

\paragraph{\textbf{Plan-aware Visual Question Answering (pVQA).}} 
This task extends the previous one to include requests that require the comprehension of user-provided images. When provided with an image $I_{k,j+1}$ and a request $u_{k,j+1}$, the model must reason about the dialogue history $T_{k,1:j}$, the instructional plan $P_k$, and the given image, to generate a text response. For this task, we extend the same objective as in PGAG, now also conditioned on the provided image $I_{k,j+1}$. The loss function is defined as:
\begin{equation}
\small
  \mathcal{L}_{\text{pVQA}} = -\sum_{t=1} \log P( w_t | T_{k,1:j}, u_{k,j+1}, I_{k,j+1}, P_k)
\end{equation}

It is important to note that, to support pVQA, the model develops a disambiguation capability that grounds its answers in the visual question rather than the instructional video.

\subsubsection{Plan-step retrieval capabilities}
The second set of model capabilities aim at retrieving an arbritrary plan step: a text step or a video moment with a text only or image query.

\paragraph{{Visually-Informed Step Generation (VSG).}} Given an unseen user-query-image, the model needs to align it with the plan to generate the corresponding action $a_j \in P_k$. This generation occurs with an in-context retrieval strategy that copies the step from the plan that is in the context of the model. The objective function is similar to the previous one:
\begin{equation}
\small
  \mathcal{L}_{\text{VSG}} = -\sum_{t=1} \log P( w_t | T_{k,1:j}, u_{k,j+1}, I_{k,j+1}, P_k)
\end{equation}

Again, the model acquires novel reasoning capabilities that allow it to navigate to a specific instructional plan action and not answer a visual question.

\paragraph{\textbf{Conversational Video Moment Retrieval.}}
The CVMR task involves retrieving a relevant video moment $vm = \langle f_{k, \text{start}}, \dots, f_{k, \text{end}} \rangle \in V_k$ that corresponds to the current plan step, addressed in the request $u_{k,j+1}$. 
We tackle this in a principled manner by retrieving the start and end frames of the relevant video moment. Specifically, we introduce two dedicated textual tokens, \texttt{[RETs]} and \texttt{[RETe]}, whose output embeddings are trained to approximate the average representation of the first and last $N$ frames of the relevant video moment, respectively.
This is optimized using a contrastive loss, specifically InfoNCE~\cite{infonce}, which tries to maximize the similarity between the \texttt{[RETs]} and \texttt{[RETe]} tokens output representations and the target representations and uses in-batch samples as negative examples. We add the PGAG loss to regularize the model to generate coherent responses to these requests. The loss function is defined as:
\begin{equation}
\small
\begin{split}
  \mathcal{L}_{\text{RET}} = \text{InfoNCE}\left( g\left(\text{\texttt{[RETs]}}\right ), \frac{1}{N} \sum_{i=0}^{N-1} v_{frame}(f_{\text{start}+i}) \right) \\ + \text{InfoNCE}\left(g\left(\text{\texttt{[RETe]}}\right ),  \frac{1}{N} \sum_{i=0}^{N-1} v_{frame}(f_{\text{end}-i}) \right) 
  \\ \mathcal{L}_{\text{CVMR}} = \frac{\mathcal{L}_{\text{RET}}}{2} + \mathcal{L}_{\text{PGAG}}
\end{split}
\end{equation} 
where $g(x)$ denotes the projected retrieval token $x$ embedding and $v_{frame}(f)$ the retrieval representation of $f$ (both defined later in Section~\ref{subsec:model_arch}).

\subsection{Model Architecture}
\label{subsec:model_arch}

\begin{table}[]
    \centering
    \small
    \resizebox{\linewidth}{!}{%
    \begin{tabular}{lc|c}
    \toprule
         & \textbf{\modelname{}} & \textbf{MM-PlanLLM} \\ \midrule
        VQA support & \textbf{\checkmark} & $\times$ \\
        Vision Tokens & \textbf{257 tokens} & 1 token \\
        CVMR Retrieval & \textbf{Start \& End Frame} & Mid Frame \\
        Multiple Domains & \textbf{\checkmark} & $\times$ \\
        \bottomrule
    \end{tabular}%
    }
    \caption{Direct comparison of \modelname{} with current procedural plan guidance SotA MM-PlanLLM~\cite{mmplanllm}} 
    \label{tab:mmplanllm_comp}
\end{table}

We design a custom multi-goal architecture (see Figure~\ref{fig:model_arch}) designed to effectively integrate multimodal information, while being flexible enough to accommodate the defined tasks:

\vspace{1mm}
\noindent\textbf{Language Model (LM) \& Vision Encoder (VE).} For the LM backbone, we adopt a pretrained LLM to leverage the knowledge and capabilities learned during pretraining and instruction tuning. We use a pretrained vision encoder, $ve(\cdot)$, to obtain region-level visual features for target retrieval video frames and uploaded images. We follow~\citet{mmstar, idefics2} and use all visual features as vision tokens. To bridge the modalities, we follow~\citet{Liu2023-vr} and connect these 2 backbones using a 2-layer projector, $W_{query}$.

\vspace{1mm}
\noindent\textbf{Retrieval Projectors.}
To deliver retrieval capabilities, two linear projection layers are learned $W_{start} \in \mathbb{R}^{d_{lm} \times d}$ and $W_{end} \in \mathbb{R}^{d_{lm} \times d}$, where $d$ is the dimension of the shared retrieval space and $d_{lm}$ is the output dimension of the LM backbone. A third layer $W_{ret} \in \mathbb{R}^{d_{f_V} \times d}$ is learned to map the visual encoder \texttt{[CLS]} token to the retrieval space. 
Following MM-PlanLLM, video temporal information is encoded by applying RoPE~\cite{rope} to the $W_{ret}$ output features, using each frame's relative position in the video. Formally, our set of generative retrieval features are learned as:
\begin{align*}
    &\text{RET Start Tok.:}\ g\left(\text{\texttt{[RETs]}}\right ) = W_{\text{start}} \cdot h\left(\text{\texttt{[RETs]}}\right) \\
    &\text{RET End Tok.:}\ \ g\left(\text{\texttt{[RETe]}}\right ) = W_{\text{end}} \cdot h\left(\text{\texttt{[RETe]}}\right)
\end{align*}
where $h$ is the last hidden layer embedding of the LM backbone. Target and candidate video frames are mapped to the same retrieval space as:
\begin{equation}
    v_{\text{frame}} = \text{RoPE} \left( W_{\text{ret}} \cdot ve\left( f_{k,pos} \right) , pos\right)
\end{equation}
here $pos$ is an integer with the position of the frame in the video and $ve(\cdot)$ is the vision encoder. 

To highlight \modelname{}'s novelty, Table~\ref{tab:mmplanllm_comp} summarizes \modelname{} key distinctions to current SotA model in procedural plan guidance, MM-PlanLLM~\cite{mmplanllm}.

\subsection{Training}

Multistage LVLM training has been the predominant training paradigm, with several works highlighting its benefit~\cite{cambrian, apollo, sharegpt4v}. \modelname{} is trained with the following structure (shown in Table~\ref{tab:training_stages}):~\footnote{For more details on each stage see Appendix~\ref{app:app_datasets}}

\vspace{1mm}
\noindent\textbf{Stage 1: Initialization.} The initial stage targets bridging the modality gap between modalities, by aligning the visual features extracted by the VE with the LM representation space, with both remaining frozen. The $W_{query}$ connector is trained on a captioning task, and the retrieval-specific projection layers, $W_{\text{ret}}, W_{\text{start}}$, and $W_{\text{end}}$, are initialized with a text-to-image retrieval task.

\vspace{1mm}
\noindent\textbf{Stage 2: Visual Instruction Tuning.} This stage is aimed at introducing strong general image understanding, in an effort to improve the final model's performance on unseen tasks. For pVQA and VSG, the model is trained end-to-end on general visual instruction datasets, whereas for retrieval captioning datasets are repurposed for image retrieval.

\vspace{1mm}
\noindent\textbf{Stage 3: Domain-specific Training.} The third stage focuses on domain specialization using datasets with in-domain dialogues or instructions to start to expose the model to the target domain. For video datasets, such as YouCook2~\cite{youcook2} and CrossTask~\cite{crosstask}, we use instructional video annotations to construct instructional training samples where the model needs to identify the plan step to which 4 provided frames belong. For retrieval, the model is trained to retrieve the relevant video moment provided a plan step and title. The FoodDialogues~\cite{foodllm} and FoodReasonSeg~\cite{foodllm} datasets are used for their dialogues with domain-specific knowledge.

\vspace{1mm}
\noindent\textbf{Stage 4: Task-specific Training.} The final stage trains the model for conversational instructional plan guidance, exposing the model to the domain-specific conversational patterns, CVMR, and VSG. 
The visual encoder $ve(\cdot)$ is frozen to avoid inserting a plan-specific bias that hinders the model's generalization capabilities, as the tasks are limited. Here, we train solely on \datasetname{}, described in Section~\ref{sec:synth_dataset}.

\subsection{\datasetname~Dataset}
\label{sec:synth_dataset}

One of our core contributions is the introduction of a conversational plan-guiding dataset that extends TastyVidDial~\cite{mmplanllm}, including both cooking and DIY plans, and adding plan-grounded visual question answering. Thus greatly widening the dataset scope and allowing for a more comprehensive evaluation on instructional plan guidance. Below we outline the details of each step of the dataset construction process:

\vspace{1mm}
\noindent\textbf{DIY Plans.}
To improve plan diversity and promote more generalizable learning, we leverage the COIN~\cite{coin} dataset to include DIY instructional plan dialogues. To do this, for each type of task in COIN, we randomly select 4 videos and then replicate the dialogue generation process from TastyVidDial. As COIN videos are not paired with instructional plans, we produce a step-by-step plan from the textual video annotations by prompting Claude 3.5~\footnote{\url{https://www.anthropic.com/news/claude-3-5-sonnet}} to rewrite the annotation sequence into a instructional plan. This results in ~700 additional dialogues.

\vspace{1mm}
\noindent\textbf{\datasetname{} Curation.}
TastyVidDial contains 1.5k illustrated recipes, with a total of 50k generated dialogues, with an average of 33 dialogues per recipe. This led to redundancy within the dataset, as many of the dialogues share the same plan. 
To mitigate this, we trim the dataset keeping only the 4 dialogues per plan that contained the most multimodal turns.

\vspace{1mm}
\noindent\textbf{Visual Question Answering Augmentation.}
To introduce pVQA requests into the already generated dialogues, we navigate the dialogues and introduce a 30\% chance of inserting a pVQA request after each time the user progresses to a new action in the plan. For each, we select the middle frame of the relevant video moment to mimic a user-uploaded image, and prompt Claude 3.5 to generate a (question, answer) pair based on the image, the provided dialogue context, and the instructional plan. This results in dialogues where pVQA requests are naturally woven into the conversation (Appendix~\ref{app:vqa_gen_details}).

\section{Experimental Setup}
\label{sec:experimental_setup}

\subsection{Multitask Training Details} 
As \modelname{} is trained in a multitask setup, we use a custom trainer, where each minibatch is sampled from multiple dataloaders to ensure a balanced task mix. Training batches are limited, per epoch, to the size of the smaller dataloader, ensuring a 1:1 sample ratio. As retrieval samples tend to have less tokens, task-specific batch sizes are used, which is beneficial for the InfoNCE retrieval loss as it uses as in-batch negatives. For retrieval, we set $d=512$, and $N=5$.~\footnote{Appendix~\ref{app:train_details} includes more details on the training setup.\label{training_footnote}}

\subsection{Datasets}
During our multistage training, the training data size progressively decreases at each stage as specialization occurs.
For stage 4, ~\datasetname{} has 6760 dialogues, which leads to a total of $\approx 114k$ dialogue turns, with a mix of multimodal requests (pVQA, VSG, CVMR) and textual plan-grounded answer generation. During training, we consider the 4 previous turns, and split the dataset in 90/5/5 for train/dev/test, with no plan overlap between splits. To reduce the frame count, for retrieval, only 1 in every 20 frames of the instructional video is kept.~\footref{training_footnote}

\subsection{LM \& Vision Backbones}
For the LM we adopt LLaMa 3 8B~\cite{llama3}, but we compare different model sizes in Appendix~\ref{app:lm_size}
For the VE, following previous findings~\cite{cambrian, apollo, idefics2, idefics3}, we use SigLIP SO400M~\cite{siglip} as the VE with an image size of 224x224 and a patch size of 14, for a total of 257 patches per image.

\subsection{Metrics}
\label{sec:metrics}
For \textbf{PGAG} we use text generation metrics: METEOR~\cite{banerjee-lavie-2005-meteor}, ROUGE-L~\cite{Lin2004ROUGEAP}, and BERTScore~\cite{Zhang2019BERTScoreET}. For \textbf{VSG} we use ROUGE-L and Exact Match, where a match is if the generated text contains the exact step text. For \textbf{pVQA} we complement ROUGE-L with Accuracy as measured by the majority vote of 3 LLMs (Claude 3.5 Sonnet~\cite{claude}, LLaMa Vision 90B~\cite{llama3}, and Pixtral Large~\cite{MistralAITeam2024AuLarge}), which are prompted with the image, dialogue context, instructional plan, and the system response, and asked to evaluate the accuracy of the response, more details in Appendix~\ref{app:vqa_eval}. 
For \textbf{CVMR}, we consider two candidate extraction approaches: \textit{``firm"} - where candidate video moments are determined by the most similar frames to the \texttt{[RETs]} and \texttt{[RETe]} token embeddings, and \textit{``adjusted"} - where the start and end frame are the frames where the similarity with the produced \texttt{[RETs]} and \texttt{[RETe]} tokens representation drops below a certain threshold, respectively.~\footnote{See Appendix~\ref{app:cvmr_threshold} for more details and examples.}
We use Recall@k, where $k \in \{1, 5\}$ with Intersection over Union (IoU) with $m \in \{0.5, 0.7\}$.

\begin{table*}[!t]
\centering
  \caption{Results on the in-domain tasks: PGAG, pVQA, and VSG. \underline{Underlined} scores are the best among the baselines. *Accuracy is measured using the majority vote of three independent LLMs (See Section \ref{sec:experimental_setup}).}
  \label{tab:pgag}
  \label{tab:vqa_vsg}
  \small
  \begin{tabular}{lccc|ccccc}
  \toprule
  & \multicolumn{5}{c}{\textbf{Plan Reasoning}} & & \multicolumn{2}{c}{\textbf{Plan-based Retrieval}} \\
  \cmidrule{2-6} \cmidrule{8-9}
  & \multicolumn{3}{c|}{\textbf{PGAG}} & \multicolumn{2}{c}{\textbf{pVQA}} & & \multicolumn{2}{c}{\textbf{VSG}}  \\
  \textbf{Model} & \textit{ROUGE-L} & \textit{METEOR} & \textit{BERTScore} & \textit{ROUGE-L} & \textit{Accuracy*} & & \textit{ROUGE-L} & \textit{EM} \\ \midrule
  Idefics2             & 37.16 & 45.09 & 67.55 & 21.22 & 44.49 & & 20.16 & 5.85  \\
  LLaVA-1.5            & 42.47 & 40.84 & 70.55 & 26.77 & 64.11 & & 16.07 & 4.21 \\
  Qwen 2.5 VL          & 31.61 & 41.51 & 63.91 & 26.39 & 75.14 & & 16.65 & 9.47 \\
  LLaVA-OV             & 41.18 & 40.85 & 70.70 & 27.02 & 77.38 & & 32.28 & 17.89 \\
  InternVL 3.5          & 41.62 & 46.20 & 69.22 & \underline{28.31} & \underline{91.40} &  & 26.30 & 14.21 \\
  Qwen 3 VL             & 44.71 & 46.31 & 71.94 & 24.72 & 91.21 & & 33.38 & 25.26 \\
  MM-PlanLLM           & \underline{58.85} & \underline{59.34} & \underline{80.03} & 7.49 & 0.37  & & \underline{44.70} & \underline{40.00} \\
  \addlinespace[0.5ex]
  \cdashline{1-9}
  \addlinespace[0.5ex]
  \modelname{}         & \textbf{75.30}   & \textbf{76.67}     & \textbf{88.72} & \textbf{33.65} & \textbf{94.02}  & & \textbf{55.66} & \textbf{57.37} \\ 
  \bottomrule
  \end{tabular}
\end{table*}

\subsection{Baselines}
Due to the limited amount of domain-specific models and models capable of performing CVMR, we complement our baseline models with general-purpose models. Specifically, we consider the following baselines:

\textbf{MM-PlanLLM}, the current in-domain SotA model. A 7B model capable of generating plan-grounded responses and retrieving video moments, but with no pVQA training. For CVMR, as it only retrieves the middle frame, we leverage their observation that the produced retrieval embedding has high similarity with the target video moment frames so we obtain candidate video moments by considering all frames adjacent to the retrieved frame with a similarity greater than 0.3.

\textbf{TRACE}~\cite{guo2025trace}, a VMR-specific model that has strong performance in the QVHighlights~\cite{Lei2021QVHighlightsDM} dataset. This is a generative model, so we can only report Recall@1 results.

We also consider six recent $\approx$7B general-purpose baselines that have shown strong performance across multiple computer vision tasks and domains: \textbf{LLaVA-1.5}~\cite{Liu2023-vr}, \textbf{Idefics2}~\cite{idefics2}, \textbf{LLaVA-OV}~\cite{llavaonevision}, \textbf{InternVL 3.5}~\cite{wang2025internvl3}, \textbf{Qwen2.5VL}~\cite{bai2025qwen25vltechnicalreport}, and \textbf{Qwen3VL}~\cite{bai2025qwen3vltechnicalreport}.

\begin{table}[]
    \centering
    \caption{Results on the CVMR task.}
    \label{tab:cvmr_results}
    \small
    \begin{tabular}{@{}c l|cccc@{}}
      \toprule
      & \multicolumn{1}{c|}{\multirow{3}{*}{\textbf{Model}}} & \multicolumn{4}{c}{\textbf{CVMR}} \\
      & \multicolumn{1}{c|}{} & \multicolumn{2}{c}{\textbf{Recall@1}} & \multicolumn{2}{c}{\textbf{Recall@5}} \\
      & \multicolumn{1}{c|}{} & \textit{m=0.5} & \textit{m=0.7} & \textit{m=0.5} & \textit{m=0.7} \\
      \midrule
      \multirow{2}{*}{\rotatebox{90}{firm}}
      & TRACE & 5.97 & 1.63 & ---   & --- \\
      & \modelname{} & 10.49 & 2.89 & 30.38 & 11.75 \\ \midrule
      \multirow{2}{*}{\rotatebox{90}{adj.}}
      & MM-PlanLLM              & 25.32 & 12.48 & 31.46 & 16.46 \\ 
      & \modelname{} & \textbf{30.74} & \textbf{15.37} & \textbf{51.18} & \textbf{26.94} \\
      \bottomrule
    \end{tabular}
\end{table}

\begin{figure}[t]
    \centering
    \caption{Comparing R@1 performance with varying values for the similarity threshold on \datasetname{}'s dev set.}
    \includegraphics[width=0.9\linewidth, trim={0mm 0mm 0mm 0mm}, clip]{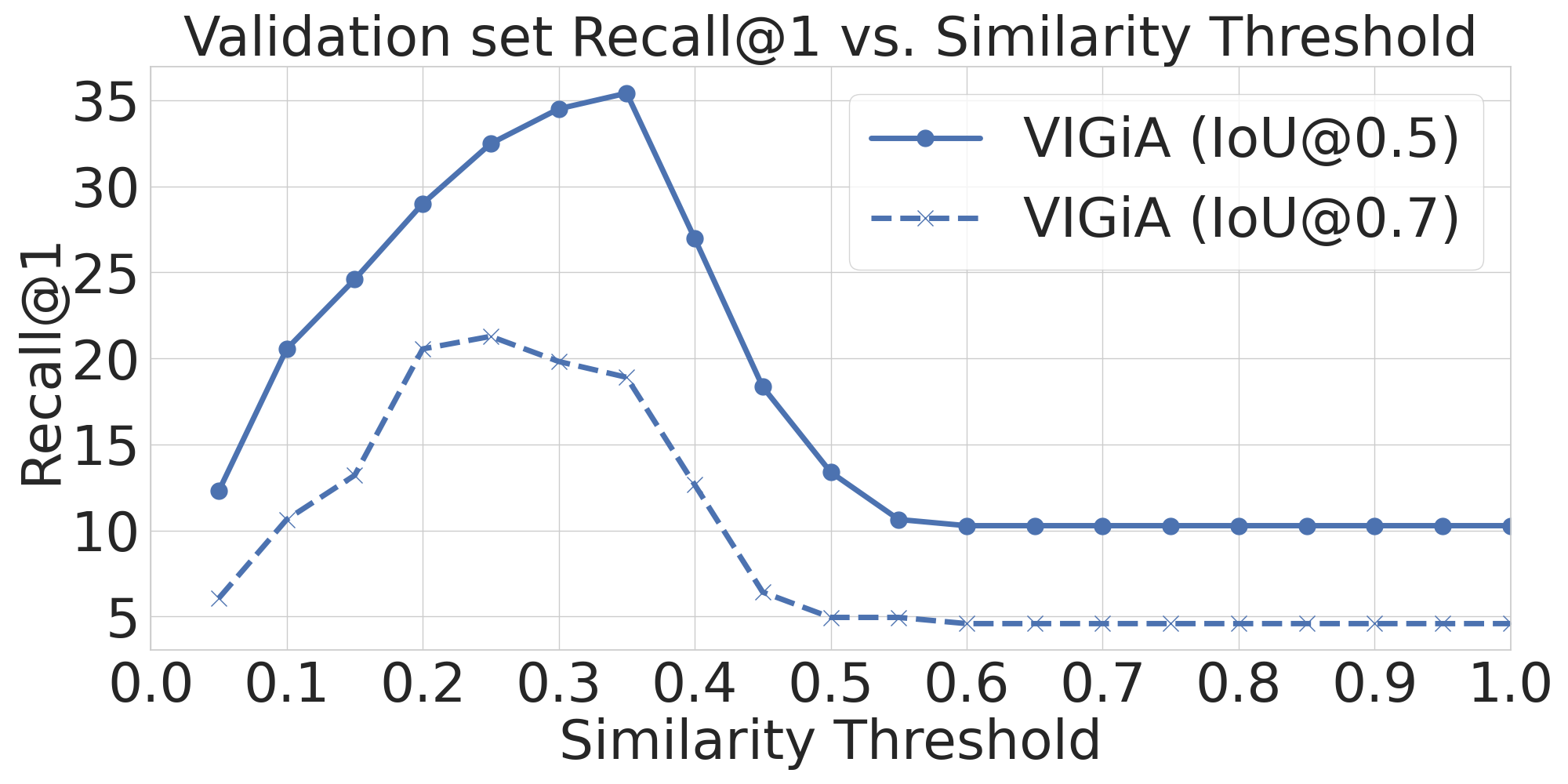}
    \label{fig:cvmr_threshold_main}
    \vspace{-2mm}
\end{figure}

\subsection{Dataset Quality Annotation}
As we add pVQA turns to the \datasetname dataset, we perform data annotation using volunteer annotators to obtain a quality estimation.
In this annotation, we asked five annotators to, once shown the complete plan, the reference image, the question, and the answer, rate (from 1 to 3) the turn on three independent measures: \textbf{Relevance} of the question to both the image and instructional plan, \textbf{Plausibility} of the question in a real setting, and \textbf{Accuracy} of the provided answer. This annotation was carried out using 250 pVQA turns. 

Our results show that the generated turns are composed of very relevant questions and accurate answers with a score of 2.67 and 2.75 respectively. Plausibility was the lowest rated measure at 2.22, although still strong. Based on annotator feedback and sample observation, we determine that the lower Plausibility score is linked to turns where, occasionally, the question is phrased as if the system is the one providing the image, not the user.~\footnote{Additional details and an error analysis are provided in Appendix~\ref{app:app_data_anno}.}

\section{Results and Discussion}
\label{sec:res_disc}

In this section, we present and discuss the results of \modelname{} on the various tasks considered. We start by evaluating on in-domain tasks using the \datasetname{} test set. This is followed by an evaluation of general-purpose image understanding using MME.

\subsection{Textual Plan-Grounded Answer Generation}
\label{sec:pgag_results}

\begin{table*}[]
\centering
\caption{Ablation of model training stages. \underline{Underlined} scores represent improvements over the previous stage/row.}
\label{tab:ablations}
\small
\begin{tabular}{@{}llcc|ccccc|ccccc@{}}
\toprule
 & & \multicolumn{4}{c}{\textbf{Plan Reasoning}} & & \multicolumn{4}{c}{\textbf{Plan-based Retrieval}} & & \multicolumn{2}{c}{\textbf{General}} \\
 \cmidrule{3-6} \cmidrule{8-11} \cmidrule{13-14}
 &  & \multicolumn{2}{c|}{\textbf{PGAG}} & \multicolumn{2}{c}{\textbf{pVQA}} & & \multicolumn{2}{c|}{\textbf{VSG}} &\multicolumn{2}{c}{\textbf{CVMR R@1}} & & \multicolumn{2}{c}{\textbf{MME}}\\
 &  & \textit{R-L} & \textit{METEOR} & \textit{R-L} & \textit{Acc.} & & \textit{R-L} & \textit{EM} & \textit{m=0.5} & \textit{m=0.7} & & \textit{Cog.} & \textit{Per.} \\ \midrule
 \multirow{4}{*}{\rotatebox{90}{Stages}} & 4 Only & 74.75 & 76.05 & 33.56 & 88.79 & & 54.27 & 56.84 & 11.39 & 5.97 & & \textbf{234} & 503 \\
 & 1+4 & 74.67 & \underline{76.61} &  \underline{33.89} & \underline{90.47} & & 54.17 & 56.32 & \underline{20.43} & \underline{11.39} & & 211 & 501 \\
 & 1+2+4 & \underline{74.99} & 76.39 & \underline{\textbf{34.20}} & \underline{91.03} & & 54.12 & \underline{56.84}& 9.58 & 5.06 & & \underline{223} & \underline{708} \\
 & 1+2+3+4 & \underline{\textbf{75.30}} & \underline{\textbf{76.67}} & 33.65 & \textbf{\underline{94.02}} & & \underline{\textbf{55.66}} & \underline{\textbf{57.37}} & \underline{\textbf{30.74}} & \underline{\textbf{15.37}} & & \underline{229} & \underline{\textbf{717}} \\ 
 \bottomrule
\end{tabular}
\end{table*}

Textual Plan-Grounded Answer Generation covers the majority of the requests seen in the \datasetname{} dataset, and is a good benchmark of the model's ability to generate coherent and contextually relevant responses based on the instructional plan dialogue. The evaluation of this task has all text-only requests from the test set, which includes a mix of questions, navigation requests, and more. The results are summarized in Table~\ref{tab:pgag}. 

These results show that \modelname{} outperforms all baselines and also reinforces the need for domain-specific training for instructional plans, as MM-PlanLLM, the only in-domain baseline, outperforms general purpose baselines.

\subsection{Multimodal Answer Generation}
\label{sec:vqa_vsg_results}

Regarding multimodal answer generation, we test the model on tasks where direct image understanding is needed to respond accurately.
Specifically, we evaluate \modelname{}'s performance, against the baselines, on the VSG and pVQA turns of the test set and present the results in Table~\ref{tab:vqa_vsg}.
The \textbf{pVQA} results attest the benefit of in-domain pVQA training as \modelname{} has a 94.02\% accuracy, a meaningful jump over most baselines, while MM-PlanLLM, which was not trained for any type of VQA, has near-zero performance, highlighting the critical gap that we are the first to address.
\textbf{VSG} results are similar with \modelname{} strongly outperforming all baselines. General purpose baselines struggle as this task demands aligning the provided image with the textual plan, requiring a strong understanding of instructional plan structure on both modalities. We also hypothesize that the inclusion of in-domain pVQA in the training data elicits richer plan understanding as we outperform MM-PlanLLM by over 25\% despite training on 5x less in-domain data.

\subsection{Conversational Video Moment Retrieval}
\label{sec:cvmr_eval}

The \textbf{CVMR} task requires retrieving a specific video moment from an instructional video, based on the dialogue context, and, as covered in Section~\ref{sec:metrics}, we evaluate two moment extraction approaches. For the \textit{``adjusted"} approach the threshold was set to 0.35, based on our observations on the validation set, shown in Figure~\ref{fig:cvmr_threshold_main}.
Table~\ref{tab:cvmr_results} shows the results of this evaluation. \modelname{} substantially improves on the previous SotA for this task, with a decent improvement over MM-PlanLLM. Comparing the \textit{``firm"} and \textit{``adjusted"} settings, we note that the latter effectively leverages the representations learned by our retrieval-specific layers and highlights the benefit of learning start/end region representations, with much higher R@1 scores. TRACE while a strong baseline for general VMR struggles in our domain-specific task likely due to the conversational setting, showing that off-the-self models are not viable for this task.

\subsection{General Task Performance}
\label{sec:gen_task_perf}

\begin{table}[]
\centering
\caption{MME performance across stages.}
\label{tab:general_mme}
\small
\begin{tabular}{@{}l|cc|cc@{}}
\toprule
\multirow{2}{*}{\textbf{Model}} & \multicolumn{2}{c|}{\textbf{Stage 2}} & \multicolumn{2}{c}{\textbf{Stage 4 / Final}} \\
\cmidrule(lr){2-3} \cmidrule(lr){4-5}
& \textit{Cog.} & \textit{Per.} & \textit{Cog.} & \textit{Per.} \\ \midrule
MM-PlanLLM       & ---   & ---   & 34 & 1   \\
Idefics2         & ---   & ---   & 328 & 1474  \\
LLaVA-1.5        & ---   & ---   & 314 & 1483  \\
LLaVA-OV         & ---   & ---   & 418 & 1580  \\
Qwen 2.5 VL      & ---   & ---   & 638 & 1685  \\
InternVL 3.5     & ---   & ---   & \textbf{663} & 1686  \\
Qwen 3 VL        & ---   & ---   & 639 & \textbf{1733}  \\ \midrule

\modelname{} & 263 & 1344 & 229 & 717  \\ 
\bottomrule
\end{tabular}
\end{table}

Domain-specific models still benefit from rich general image understanding to better address unseen plans and domains.
We use MME~\cite{fu2023mme}, a benchmark of 14 image understanding sub-tasks, as a proxy to measure flexibility and the impact of domain-specific training.
 
From Table~\ref{tab:general_mme} we see the expected performance plot with general purpose models excelling. Regarding \modelname{}, we confirm that Stage 2 training instills rich image understanding but, after Stage 4, there is a performance drop, a trade-off of as the model begins to be specialized.
In general, \modelname{} retains general image understanding capabilities even after instructional plan guidance specialization, particularly compared to MM-PlanLLM.

\subsection{Dialogue-level Evaluation}
\label{sec:dialogue_level_eval}

\begin{table}[]
\centering
\caption{Dialogue-level evaluation ($N=54$) with LLM-as-a-judge to measure guidance quality.}
\label{tab:dialogue_level}
\small
\begin{tabular}{@{}l|ccc@{}}
\toprule
 \multirow{2}{*}{\textbf{Model}} & \textbf{State} & \textbf{Instruction} & \textbf{Plan} \\
  & \textbf{Tracking} & \textbf{Clarity} & \textbf{Adherence} \\ \midrule
 MM-PlanLLM     & 1.61 & 2.11 & 2.24 \\
 Qwen 2.5 VL    & 2.31 & 2.41 & 2.74 \\
 Llava OV       & 1.87 & 2.59 & 3.00 \\
 InternVL 3.5   & 2.52 & 2.44 & 3.13 \\
 Qwen 3 VL      & 2.85 & 2.35 & 3.39 \\
\addlinespace[0.5ex]
\cdashline{1-4}
\addlinespace[0.5ex]
 \modelname{} & \textbf{3.30} & \textbf{4.06} & \textbf{4.26} \\
 \bottomrule
\end{tabular}
\end{table}

Our evaluation follows previous works' protocol where the evaluation is conducted at turn level. While this is useful to measure model performance on each task, it fails to capture model performance on extended dialogue reasoning. To address this gap, we collected all test set dialogues without CVMR turns (54 dialogues) and evaluated dialogue level performance. We generated full dialogue trajectories by appending model responses to the context turn-by-turn, using the original user inputs as a static simulator. This exposes the model to error propagation.
For evaluation we use an LLM-as-a-judge approach and judge each dialogue on 3 dimensions (in a 1-5 scale): State Tracking (accuracy of tracking the user progress across the conversation), Instruction Clarity (absence of added unwanted commentary that hinders instruction comprehension), and Plan Adherence (faithfulness to the plan, not hallucinating any steps or ingredients/tools). More details on this evaluation setup are provided in Appendix~\ref{app:dialogue_level}.

Table~\ref{tab:dialogue_level} shows the results of this evaluation. Here we see that \modelname{} performs strongly on all three dimensions with a significant lead in Instruction Clarity. Qualitatively, baseline models tend to be extremely verbose, often adding markdown notation and additional commentary, resulting in poor Instruction Clarity, an important factor when guiding user unfamiliar with the task. 
Finally, State Tracking emerges as the most challenging dimension, reinforcing the difficulty of tracking user progress through both visual and textual inputs over extend dialogue sessions where progress is rarely linear, one of the core challenges of procedural plan guidance.

\subsection{Ablation Studies}

\begin{table}[]
\centering
\caption{Ablation different pVQA data amounts.}
\label{tab:pvqa_ablations}
\small
\begin{tabular}{@{}lcc|cc|cc@{}}
\toprule
 \textbf{pVQA} & \multicolumn{2}{c|}{\textbf{PGAG}} & \multicolumn{2}{c|}{\textbf{pVQA}} & \multicolumn{2}{c}{\textbf{VSG}} \\
 \textbf{Data \%} & \textit{R-L} & \textit{M} & \textit{R-L} & \textit{Acc.} & \textit{R-L} & \textit{EM} \\ \midrule
 0\% & 75.62 & \textbf{77.03} & 23.30 & 71.59 & 53.54 & 56.32\\
 10\% & 75.63 & 76.98 & 31.66 & 85.23 & 53.55 & 54.74 \\
 20\% & 75.56 & 76.97 & 32.41 & 85.42 & 53.17 & 55.79 \\
 50\% & \textbf{76.09} & 76.54 & \textbf{34.00} & 88.79 & \textbf{56.01} & \textbf{57.89} \\
 100\% & 75.30 & 76.67 & 33.65 & \textbf{94.02} & 55.66 & 57.37
 \\
 \bottomrule
\end{tabular}
\end{table}

To evaluate and measure the impact of each training stage on the different tasks, we performed an ablation study by progressively adding additional training stages.
Table~\ref{tab:ablations} shows our ablation of multistage training. We see an incremental benefit of each added training stage. Although the single-stage approach presents a strong baseline, it lacks CVMR and MME performance, which are then addressed in the latter stages. The first and third stage collectively improve CVMR performance, up 20 points on R@1 with m=0.5. The 2nd stage improves model flexibility with an MME Perception increase of 200 points, without a significant loss on in-domain tasks. We also ablated RoPE usage for frame position encoding and found that not using it degraded performance, with R@1 m=0.5 dropping 13\%, from 30.74 to 26.76 and R@1 m=0.7 dropping 29\% from 15.37 to 11.03.

Additionally, we evaluated the impact of pVQA data volume in training. From Table~\ref{tab:pvqa_ablations}, we highlight two key findings: 1) including pVQA data yielded a notable improvement in VSG performance, likely due to their shared MLP module; and 2) performance across most metrics peaks with 50\% pVQA data, with a minor decrease when all pVQA samples are used, likely due to task saturation.
A further ablation of a text-only version of \modelname{} is included in Appendix~\ref{app:text_only}.

\section{Conclusion}

This paper proposes \modelname{}, the first LVLM model for conversational instructional plan guidance, that jointly addresses the key goals of the task: 1) rich image understanding for Plan-grounded pVQA, 2) conversational video moment retrieval, 3) multimodal plan alignment and 4) plan-grounded answer generation. An extended cross-modal architecture is designed, supporting unified plan-grounded dialogue and generative video moment retrieval.

In our detailed evaluation, we compare \modelname{} against both domain-specific and general-purpose LVLMs and find that it consistently and significantly outperforms both types of baselines, with over 90\% accuracy in plan-grounded in-context VQA. We complement our analysis with an evaluation on a general image understanding benchmark and show that \modelname{} remains far more capable than the in-domain state-of-the-art.

\section{Limitations}

The very limited amount of instructional plans paired with an annotated instructional video lead to a dataset with not enough training examples diverse enough to perform a full training, without needing auxiliary datasets. As presented in Table 5, \modelname{} is trained with different datasets over different phases. These are intended to equip the model with increasingly complex capabilities, to mitigate the knowledge gaps of the latter stages.

The datasets and instructional plans used for domain-specific training (Stage 3 \& 4) might have a western bias, as they were sourced from English sources. Because of this, we posit that for other cultural contexts model performance can vary as the model is not as familiar with region-specific terminology and techniques. However, we do not believe that core functionalities, such as navigation and context-dependent QA, are affected because these functionalities depend on task structure rather than external knowledge.

\section*{Acknowledgements}
This work was partially supported by the FCT Ph.D. scholarship grant Ref. PRT/BD/152810/2021 awarded by the CMU Portugal Affiliated Ph.D. program, by the FCT project NOVA LINCS Ref. (UIDB/04516/2020), by the Alexa Prize Taskbot Challenge organized by Amazon Science, by the AMALIA project inserted in measure RE-C05-i08 of the Portuguese national ``Programa de Recuperação e Resiliência" and by the FCT project Ref. 2024.07383.IACDC for public administration.

\bibliography{main}

@String(CVPR= {IEEE Conf. Comput. Vis. Pattern Recog.})

@String(ICCV= {Int. Conf. Comput. Vis.})

@String(ECCV= {Eur. Conf. Comput. Vis.})

@String(ICLR = {Int. Conf. Learn. Represent.})

@String(AAAI = {AAAI})

@String(CVPR  = {CVPR})

@String(ICCV  = {ICCV})

@String(ECCV  = {ECCV})

@String(ICLR  = {ICLR})

@inproceedings{Liu2023-vr,
  author       = {Haotian Liu and
                  Chunyuan Li and
                  Yuheng Li and
                  Yong Jae Lee},
  title        = {Improved Baselines with Visual Instruction Tuning},
  booktitle    = {{IEEE/CVF} Conference on Computer Vision and Pattern Recognition,
                  {CVPR} 2024, Seattle, WA, USA, June 16-22, 2024},
  pages        = {26286--26296},
  publisher    = {{IEEE}},
  year         = {2024},
  url          = {https://doi.org/10.1109/CVPR52733.2024.02484},
  doi          = {10.1109/CVPR52733.2024.02484},
  bibsource    = {dblp computer science bibliography, https://dblp.org}
}

@inproceedings{llava,
  author       = {Haotian Liu and
                  Chunyuan Li and
                  Qingyang Wu and
                  Yong Jae Lee},
  editor       = {Alice Oh and
                  Tristan Naumann and
                  Amir Globerson and
                  Kate Saenko and
                  Moritz Hardt and
                  Sergey Levine},
  title        = {Visual Instruction Tuning},
  booktitle    = {Advances in Neural Information Processing Systems 36: Annual Conference
                  on Neural Information Processing Systems 2023, NeurIPS 2023, New Orleans,
                  LA, USA, December 10 - 16, 2023},
  year         = {2023},
  url          = {http://papers.nips.cc/paper\_files/paper/2023/hash/6dcf277ea32ce3288914faf369fe6de0-Abstract-Conference.html},
  bibsource    = {dblp computer science bibliography, https://dblp.org}
}

@inproceedings{llava_rlhf,
  author       = {Zhiqing Sun and
                  Sheng Shen and
                  Shengcao Cao and
                  Haotian Liu and
                  Chunyuan Li and
                  Yikang Shen and
                  Chuang Gan and
                  Liangyan Gui and
                  Yu{-}Xiong Wang and
                  Yiming Yang and
                  Kurt Keutzer and
                  Trevor Darrell},
  editor       = {Lun{-}Wei Ku and
                  Andre Martins and
                  Vivek Srikumar},
  title        = {Aligning Large Multimodal Models with Factually Augmented {RLHF}},
  booktitle    = {Findings of the Association for Computational Linguistics, {ACL} 2024,
                  Bangkok, Thailand and virtual meeting, August 11-16, 2024},
  pages        = {13088--13110},
  publisher    = {Association for Computational Linguistics},
  year         = {2024},
  url          = {https://doi.org/10.18653/v1/2024.findings-acl.775},
  doi          = {10.18653/V1/2024.FINDINGS-ACL.775},
  bibsource    = {dblp computer science bibliography, https://dblp.org}
}

@ARTICLE{llavar,
  title         = "{LLaVAR}: Enhanced Visual Instruction Tuning for {Text-Rich}
                   Image Understanding",
  author        = "Zhang, Yanzhe and Zhang, Ruiyi and Gu, Jiuxiang and Zhou,
                   Yufan and Lipka, Nedim and Yang, Diyi and Sun, Tong",
  month         =  jun,
  year          =  2023,
  archivePrefix = "arXiv",
  primaryClass  = "cs.CV",
  eprint        = "2306.17107"
}

@inproceedings{flamingo,
 author = {Alayrac, Jean-Baptiste and Donahue, Jeff and Luc, Pauline and Miech, Antoine and Barr, Iain and Hasson, Yana and Lenc, Karel and Mensch, Arthur and Millican, Katherine and Reynolds, Malcolm and Ring, Roman and Rutherford, Eliza and Cabi, Serkan and Han, Tengda and Gong, Zhitao and Samangooei, Sina and Monteiro, Marianne and Menick, Jacob L and Borgeaud, Sebastian and Brock, Andy and Nematzadeh, Aida and Sharifzadeh, Sahand and Bi\'{n}kowski, Miko\l aj and Barreira, Ricardo and Vinyals, Oriol and Zisserman, Andrew and Simonyan, Kar\'{e}n},
 booktitle = {Advances in Neural Information Processing Systems},
 editor = {S. Koyejo and S. Mohamed and A. Agarwal and D. Belgrave and K. Cho and A. Oh},
 pages = {23716--23736},
 publisher = {Curran Associates, Inc.},
 title = {Flamingo: a Visual Language Model for Few-Shot Learning},
 url = {https://proceedings.neurips.cc/paper_files/paper/2022/file/960a172bc7fbf0177ccccbb411a7d800-Paper-Conference.pdf},
 volume = {35},
 year = {2022}
}

@inproceedings{blip_2,
  author       = {Junnan Li and
                  Dongxu Li and
                  Silvio Savarese and
                  Steven C. H. Hoi},
  editor       = {Andreas Krause and
                  Emma Brunskill and
                  Kyunghyun Cho and
                  Barbara Engelhardt and
                  Sivan Sabato and
                  Jonathan Scarlett},
  title        = {{BLIP-2:} Bootstrapping Language-Image Pre-training with Frozen Image
                  Encoders and Large Language Models},
  booktitle    = {International Conference on Machine Learning, {ICML} 2023, 23-29 July
                  2023, Honolulu, Hawaii, {USA}},
  series       = {Proceedings of Machine Learning Research},
  volume       = {202},
  pages        = {19730--19742},
  publisher    = {{PMLR}},
  year         = {2023},
  url          = {https://proceedings.mlr.press/v202/li23q.html},
  bibsource    = {dblp computer science bibliography, https://dblp.org}
}

@inproceedings{instructblip,
  author       = {Wenliang Dai and
                  Junnan Li and
                  Dongxu Li and
                  Anthony Meng Huat Tiong and
                  Junqi Zhao and
                  Weisheng Wang and
                  Boyang Li and
                  Pascale Fung and
                  Steven C. H. Hoi},
  editor       = {Alice Oh and
                  Tristan Naumann and
                  Amir Globerson and
                  Kate Saenko and
                  Moritz Hardt and
                  Sergey Levine},
  title        = {InstructBLIP: Towards General-purpose Vision-Language Models with
                  Instruction Tuning},
  booktitle    = {Advances in Neural Information Processing Systems 36: Annual Conference
                  on Neural Information Processing Systems 2023, NeurIPS 2023, New Orleans,
                  LA, USA, December 10 - 16, 2023},
  year         = {2023},
  url          = {http://papers.nips.cc/paper\_files/paper/2023/hash/9a6a435e75419a836fe47ab6793623e6-Abstract-Conference.html},
  bibsource    = {dblp computer science bibliography, https://dblp.org}
}

@inproceedings{planllm,
    title = "Plan-Grounded Large Language Models for Dual Goal Conversational Settings",
    author = "Gl{\'o}ria-Silva, Diogo  and
      Ferreira, Rafael  and
      Tavares, Diogo  and
      Semedo, David  and
      Magalhaes, Joao",
    editor = "Graham, Yvette  and
      Purver, Matthew",
    booktitle = "Proceedings of the 18th Conference of the European Chapter of the Association for Computational Linguistics (Volume 1: Long Papers)",
    month = mar,
    year = "2024",
    address = "St. Julian{'}s, Malta",
    publisher = "Association for Computational Linguistics",
    url = "https://aclanthology.org/2024.eacl-long.77",
    pages = "1271--1292",
}

@article{llama3,
  author       = {Abhimanyu Dubey and
                  Abhinav Jauhri and
                  Abhinav Pandey and
                  Abhishek Kadian and
                  Ahmad Al{-}Dahle and
                  Aiesha Letman and
                  Akhil Mathur and
                  Alan Schelten and
                  Amy Yang and
                  Angela Fan and
                  Anirudh Goyal and
                  Anthony Hartshorn and
                  Aobo Yang and
                  Archi Mitra and
                  Archie Sravankumar and
                  Artem Korenev and
                  Arthur Hinsvark and
                  Arun Rao and
                  Aston Zhang and
                  Aur{\'{e}}lien Rodriguez and
                  Austen Gregerson and
                  Ava Spataru and
                  Baptiste Rozi{\`{e}}re and
                  Bethany Biron and
                  Binh Tang and
                  Bobbie Chern and
                  Charlotte Caucheteux and
                  Chaya Nayak and
                  Chloe Bi and
                  Chris Marra and
                  Chris McConnell and
                  Christian Keller and
                  Christophe Touret and
                  Chunyang Wu and
                  Corinne Wong and
                  Cristian Canton Ferrer and
                  Cyrus Nikolaidis and
                  Damien Allonsius and
                  Daniel Song and
                  Danielle Pintz and
                  Danny Livshits and
                  David Esiobu and
                  Dhruv Choudhary and
                  Dhruv Mahajan and
                  Diego Garcia{-}Olano and
                  Diego Perino and
                  Dieuwke Hupkes and
                  Egor Lakomkin and
                  Ehab AlBadawy and
                  Elina Lobanova and
                  Emily Dinan and
                  Eric Michael Smith and
                  Filip Radenovic and
                  Frank Zhang and
                  Gabriel Synnaeve and
                  Gabrielle Lee and
                  Georgia Lewis Anderson and
                  Graeme Nail and
                  Gr{\'{e}}goire Mialon and
                  Guan Pang and
                  Guillem Cucurell and
                  Hailey Nguyen and
                  Hannah Korevaar and
                  Hu Xu and
                  Hugo Touvron and
                  Iliyan Zarov and
                  Imanol Arrieta Ibarra and
                  Isabel M. Kloumann and
                  Ishan Misra and
                  Ivan Evtimov and
                  Jade Copet and
                  Jaewon Lee and
                  Jan Geffert and
                  Jana Vranes and
                  Jason Park and
                  Jay Mahadeokar and
                  Jeet Shah and
                  Jelmer van der Linde and
                  Jennifer Billock and
                  Jenny Hong and
                  Jenya Lee and
                  Jeremy Fu and
                  Jianfeng Chi and
                  Jianyu Huang and
                  Jiawen Liu and
                  Jie Wang and
                  Jiecao Yu and
                  Joanna Bitton and
                  Joe Spisak and
                  Jongsoo Park and
                  Joseph Rocca and
                  Joshua Johnstun and
                  Joshua Saxe and
                  Junteng Jia and
                  Kalyan Vasuden Alwala and
                  Kartikeya Upasani and
                  Kate Plawiak and
                  Ke Li and
                  Kenneth Heafield and
                  Kevin Stone and
                  et al.},
  title        = {The Llama 3 Herd of Models},
  journal      = {CoRR},
  volume       = {abs/2407.21783},
  year         = {2024},
  url          = {https://doi.org/10.48550/arXiv.2407.21783},
  doi          = {10.48550/ARXIV.2407.21783},
  eprinttype    = {arXiv},
  eprint       = {2407.21783},
  bibsource    = {dblp computer science bibliography, https://dblp.org}
}

@INPROCEEDINGS{tasty_dataset,
  author={Sener, Fadime and Yao, Angela},
  booktitle={2019 IEEE/CVF International Conference on Computer Vision (ICCV)}, 
  title={Zero-Shot Anticipation for Instructional Activities}, 
  year={2019},
  volume={},
  number={},
  pages={862-871},
  doi={10.1109/ICCV.2019.00095}
}

@inproceedings{Lin2004ROUGEAP,
    title = "{ROUGE}: A Package for Automatic Evaluation of Summaries",
    author = "Lin, Chin-Yew",
    booktitle = "Text Summarization Branches Out",
    month = jul,
    year = "2004",
    address = "Barcelona, Spain",
    publisher = "Association for Computational Linguistics",
    url = "https://aclanthology.org/W04-1013/",
    pages = "74--81"
}

@inproceedings{Zhang2019BERTScoreET,
  author       = {Tianyi Zhang and
                  Varsha Kishore and
                  Felix Wu and
                  Kilian Q. Weinberger and
                  Yoav Artzi},
  title        = {BERTScore: Evaluating Text Generation with {BERT}},
  booktitle    = {8th International Conference on Learning Representations, {ICLR} 2020,
                  Addis Ababa, Ethiopia, April 26-30, 2020},
  publisher    = {OpenReview.net},
  year         = {2020},
  url          = {https://openreview.net/forum?id=SkeHuCVFDr},
  bibsource    = {dblp computer science bibliography, https://dblp.org}
}

@inproceedings{wizard_of_tasks,
  author    = {Jason Ingyu Choi and
               Saar Kuzi and
               Nikhita Vedula and
               Jie Zhao and
               Giuseppe Castellucci and
               Marcus Collins and
               Shervin Malmasi and
               Oleg Rokhlenko and
               Eugene Agichtein},
  title     = {Wizard of Tasks: {A} Novel Conversational Dataset for Solving Real-World
               Tasks in Conversational Settings},
  booktitle = {Proceedings of the 29th International Conference on Computational
               Linguistics, {COLING} 2022, Gyeongju, Republic of Korea, October 12-17,
               2022},
  pages     = {3514--3529},
  publisher = {International Committee on Computational Linguistics},
  year      = {2022},
  url       = {https://aclanthology.org/2022.coling-1.310},
  bibsource = {dblp computer science bibliography, https://dblp.org}
}

@inproceedings{perceiver,
  author       = {Andrew Jaegle and
                  Felix Gimeno and
                  Andy Brock and
                  Oriol Vinyals and
                  Andrew Zisserman and
                  Jo{\~{a}}o Carreira},
  editor       = {Marina Meila and
                  Tong Zhang},
  title        = {Perceiver: General Perception with Iterative Attention},
  booktitle    = {Proceedings of the 38th International Conference on Machine Learning,
                  {ICML} 2021, 18-24 July 2021, Virtual Event},
  series       = {Proceedings of Machine Learning Research},
  volume       = {139},
  pages        = {4651--4664},
  publisher    = {{PMLR}},
  year         = {2021},
  url          = {http://proceedings.mlr.press/v139/jaegle21a.html},
  bibsource    = {dblp computer science bibliography, https://dblp.org}
}

@article{Lei2021QVHighlightsDM,
  title={Detecting moments and highlights in videos via natural language queries},
  author={Lei, Jie and Berg, Tamara L and Bansal, Mohit},
  journal={Advances in Neural Information Processing Systems},
  volume={34},
  pages={11846--11858},
  year={2021}
}

@article{infonce,
  author       = {A{\"{a}}ron van den Oord and
                  Yazhe Li and
                  Oriol Vinyals},
  title        = {Representation Learning with Contrastive Predictive Coding},
  journal      = {CoRR},
  volume       = {abs/1807.03748},
  year         = {2018},
  url          = {http://arxiv.org/abs/1807.03748},
  eprinttype    = {arXiv},
  eprint       = {1807.03748},
  bibsource    = {dblp computer science bibliography, https://dblp.org}
}

@article{gemma,
	title={Gemma},
	url={https://www.kaggle.com/m/3301},
	DOI={10.34740/KAGGLE/M/3301},
	publisher={Kaggle},
	author={Gemma Team and Thomas Mesnard and Cassidy Hardin and Robert Dadashi and Surya Bhupatiraju and Laurent Sifre and Morgane Rivière and Mihir Sanjay Kale and Juliette Love and Pouya Tafti and et al.},
	year={2024}
}

@article{yang2024qwen2,
  title={Qwen2. 5 technical report},
  author={Yang, An and Yang, Baosong and Zhang, Beichen and Hui, Binyuan and Zheng, Bo and Yu, Bowen and Li, Chengyuan and Liu, Dayiheng and Huang, Fei and Wei, Haoran and others},
  journal={arXiv preprint arXiv:2412.15115},
  year={2024}
}

@misc{deepseekai2024deepseekv3technicalreport,
      title={DeepSeek-V3 Technical Report}, 
      author={DeepSeek-AI},
      year={2024},
      eprint={2412.19437},
      archivePrefix={arXiv},
      primaryClass={cs.CL},
      url={https://arxiv.org/abs/2412.19437}, 
}

@inproceedings{clip,
  author       = {Alec Radford and
                  Jong Wook Kim and
                  Chris Hallacy and
                  Aditya Ramesh and
                  Gabriel Goh and
                  Sandhini Agarwal and
                  Girish Sastry and
                  Amanda Askell and
                  Pamela Mishkin and
                  Jack Clark and
                  Gretchen Krueger and
                  Ilya Sutskever},
  editor       = {Marina Meila and
                  Tong Zhang},
  title        = {Learning Transferable Visual Models From Natural Language Supervision},
  booktitle    = {Proceedings of the 38th International Conference on Machine Learning,
                  {ICML} 2021, 18-24 July 2021, Virtual Event},
  series       = {Proceedings of Machine Learning Research},
  volume       = {139},
  pages        = {8748--8763},
  publisher    = {{PMLR}},
  year         = {2021},
  url          = {http://proceedings.mlr.press/v139/radford21a.html},
  bibsource    = {dblp computer science bibliography, https://dblp.org}
}

@inproceedings{
hu2022lora,
title={Lo{RA}: Low-Rank Adaptation of Large Language Models},
author={Edward J Hu and Yelong Shen and Phillip Wallis and Zeyuan Allen-Zhu and Yuanzhi Li and Shean Wang and Lu Wang and Weizhu Chen},
booktitle={International Conference on Learning Representations},
year={2022},
url={https://openreview.net/forum?id=nZeVKeeFYf9}
}

@inproceedings{fischer2024grillbot,
  author       = {Sophie Fischer and
                  Carlos Gemmell and
                  Niklas Tecklenburg and
                  Iain Mackie and
                  Federico Rossetto and
                  Jeffrey Dalton},
  editor       = {Ricardo Baeza{-}Yates and
                  Francesco Bonchi},
  title        = {GRILLBot In Practice: Lessons and Tradeoffs Deploying Large Language
                  Models for Adaptable Conversational Task Assistants},
  booktitle    = {Proceedings of the 30th {ACM} {SIGKDD} Conference on Knowledge Discovery
                  and Data Mining, {KDD} 2024, Barcelona, Spain, August 25-29, 2024},
  pages        = {4951--4961},
  publisher    = {{ACM}},
  year         = {2024},
  url          = {https://doi.org/10.1145/3637528.3671622},
  doi          = {10.1145/3637528.3671622},
  bibsource    = {dblp computer science bibliography, https://dblp.org}
}

@inproceedings{mo-etal-2023-roll,
    title = "Roll Up Your Sleeves: Working with a Collaborative and Engaging Task-Oriented Dialogue System",
    author = "Mo, Lingbo  and
      Chen, Shijie  and
      Chen, Ziru  and
      Deng, Xiang  and
      Lewis, Ashley  and
      Singh, Sunit  and
      Stevens, Samuel  and
      Tai, Chang-You  and
      Wang, Zhen  and
      Yue, Xiang  and
      Zhang, Tianshu  and
      Su, Yu  and
      Sun, Huan",
    editor = "Stoyanchev, Svetlana  and
      Joty, Shafiq  and
      Schlangen, David  and
      Dusek, Ondrej  and
      Kennington, Casey  and
      Alikhani, Malihe",
    booktitle = "Proceedings of the 24th Annual Meeting of the Special Interest Group on Discourse and Dialogue",
    month = sep,
    year = "2023",
    address = "Prague, Czechia",
    publisher = "Association for Computational Linguistics",
    url = "https://aclanthology.org/2023.sigdial-1.19/",
    doi = "10.18653/v1/2023.sigdial-1.19",
    pages = "197--201"
}

@article{rope,
  author       = {Jianlin Su and
                  Murtadha H. M. Ahmed and
                  Yu Lu and
                  Shengfeng Pan and
                  Wen Bo and
                  Yunfeng Liu},
  title        = {RoFormer: Enhanced transformer with Rotary Position Embedding},
  journal      = {Neurocomputing},
  volume       = {568},
  pages        = {127063},
  year         = {2024},
  url          = {https://doi.org/10.1016/j.neucom.2023.127063},
  doi          = {10.1016/J.NEUCOM.2023.127063},
  bibsource    = {dblp computer science bibliography, https://dblp.org}
}

@inproceedings{mmplanllm,
    title = "Show and Guide: Instructional-Plan Grounded Vision and Language Model",
    author = "Gl{\'o}ria-Silva, Diogo  and
      Semedo, David  and
      Magalhaes, Joao",
    editor = "Al-Onaizan, Yaser  and
      Bansal, Mohit  and
      Chen, Yun-Nung",
    booktitle = "Proceedings of the 2024 Conference on Empirical Methods in Natural Language Processing",
    month = nov,
    year = "2024",
    address = "Miami, Florida, USA",
    publisher = "Association for Computational Linguistics",
    url = "https://aclanthology.org/2024.emnlp-main.1191/",
    doi = "10.18653/v1/2024.emnlp-main.1191",
    pages = "21371--21389"
}

@article{idefics2,
  title={What matters when building vision-language models?},
  author={Lauren{\c{c}}on, Hugo and Tronchon, L{\'e}o and Cord, Matthieu and Sanh, Victor},
  journal={Advances in Neural Information Processing Systems},
  volume={37},
  pages={87874--87907},
  year={2024}
}

@inproceedings{cambrian,
 author = {Tong, Shengbang and Brown, Ellis and Wu, Penghao and Woo, Sanghyun and Middepogu, Manoj and Akula, Sai Charitha and Yang, Jihan and Yang, Shusheng and Iyer, Adithya and Pan, Xichen and Wang, Austin and Fergus, Rob and LeCun, Yann and Xie, Saining},
 booktitle = {Advances in Neural Information Processing Systems},
 editor = {A. Globerson and L. Mackey and D. Belgrave and A. Fan and U. Paquet and J. Tomczak and C. Zhang},
 pages = {87310--87356},
 publisher = {Curran Associates, Inc.},
 title = {Cambrian-1: A Fully Open, Vision-Centric Exploration of Multimodal LLMs},
 url = {https://proceedings.neurips.cc/paper_files/paper/2024/file/9ee3a664ccfeabc0da16ac6f1f1cfe59-Paper-Conference.pdf},
 volume = {37},
 year = {2024}
}

@inproceedings{mmstar,
  author       = {Lin Chen and
                  Jinsong Li and
                  Xiaoyi Dong and
                  Pan Zhang and
                  Yuhang Zang and
                  Zehui Chen and
                  Haodong Duan and
                  Jiaqi Wang and
                  Yu Qiao and
                  Dahua Lin and
                  Feng Zhao},
  editor       = {Amir Globersons and
                  Lester Mackey and
                  Danielle Belgrave and
                  Angela Fan and
                  Ulrich Paquet and
                  Jakub M. Tomczak and
                  Cheng Zhang},
  title        = {Are We on the Right Way for Evaluating Large Vision-Language Models?},
  booktitle    = {Advances in Neural Information Processing Systems 38: Annual Conference
                  on Neural Information Processing Systems 2024, NeurIPS 2024, Vancouver,
                  BC, Canada, December 10 - 15, 2024},
  year         = {2024},
  url          = {http://papers.nips.cc/paper\_files/paper/2024/hash/2f8ee6a3d766b426d2618e555b5aeb39-Abstract-Conference.html},
  bibsource    = {dblp computer science bibliography, https://dblp.org}
}

@article{apollo,
  author       = {Orr Zohar and
                  Xiaohan Wang and
                  Yann Dubois and
                  Nikhil Mehta and
                  Tong Xiao and
                  Philippe Hansen{-}Estruch and
                  Licheng Yu and
                  Xiaofang Wang and
                  Felix Juefei{-}Xu and
                  Ning Zhang and
                  Serena Yeung{-}Levy and
                  Xide Xia},
  title        = {Apollo: An Exploration of Video Understanding in Large Multimodal
                  Models},
  journal      = {CoRR},
  volume       = {abs/2412.10360},
  year         = {2024},
  url          = {https://doi.org/10.48550/arXiv.2412.10360},
  doi          = {10.48550/ARXIV.2412.10360},
  eprinttype    = {arXiv},
  eprint       = {2412.10360},
  bibsource    = {dblp computer science bibliography, https://dblp.org}
}

@inproceedings{ShareGPT4V,
  author       = {Lin Chen and
                  Jinsong Li and
                  Xiaoyi Dong and
                  Pan Zhang and
                  Conghui He and
                  Jiaqi Wang and
                  Feng Zhao and
                  Dahua Lin},
  editor       = {Ales Leonardis and
                  Elisa Ricci and
                  Stefan Roth and
                  Olga Russakovsky and
                  Torsten Sattler and
                  G{\"{u}}l Varol},
  title        = {ShareGPT4V: Improving Large Multi-modal Models with Better Captions},
  booktitle    = {Computer Vision - {ECCV} 2024 - 18th European Conference, Milan, Italy,
                  September 29-October 4, 2024, Proceedings, Part {XVII}},
  series       = {Lecture Notes in Computer Science},
  volume       = {15075},
  pages        = {370--387},
  publisher    = {Springer},
  year         = {2024},
  url          = {https://doi.org/10.1007/978-3-031-72643-9\_22},
  doi          = {10.1007/978-3-031-72643-9\_22},
  bibsource    = {dblp computer science bibliography, https://dblp.org}
}

@article{laion400m,
  author       = {Christoph Schuhmann and
                  Richard Vencu and
                  Romain Beaumont and
                  Robert Kaczmarczyk and
                  Clayton Mullis and
                  Aarush Katta and
                  Theo Coombes and
                  Jenia Jitsev and
                  Aran Komatsuzaki},
  title        = {{LAION-400M:} Open Dataset of CLIP-Filtered 400 Million Image-Text
                  Pairs},
  journal      = {CoRR},
  volume       = {abs/2111.02114},
  year         = {2021},
  url          = {https://arxiv.org/abs/2111.02114},
  eprinttype    = {arXiv},
  eprint       = {2111.02114},
  bibsource    = {dblp computer science bibliography, https://dblp.org}
}

@article{pixmo,
  author       = {Matt Deitke and
                  Christopher Clark and
                  Sangho Lee and
                  Rohun Tripathi and
                  Yue Yang and
                  Jae Sung Park and
                  Mohammadreza Salehi and
                  Niklas Muennighoff and
                  Kyle Lo and
                  Luca Soldaini and
                  Jiasen Lu and
                  Taira Anderson and
                  Erin Bransom and
                  Kiana Ehsani and
                  Huong Ngo and
                  Yen{-}Sung Chen and
                  Ajay Patel and
                  Mark Yatskar and
                  Chris Callison{-}Burch and
                  Andrew Head and
                  Rose Hendrix and
                  Favyen Bastani and
                  Eli VanderBilt and
                  Nathan Lambert and
                  Yvonne Chou and
                  Arnavi Chheda and
                  Jenna Sparks and
                  Sam Skjonsberg and
                  Michael Schmitz and
                  Aaron Sarnat and
                  Byron Bischoff and
                  Pete Walsh and
                  Chris Newell and
                  Piper Wolters and
                  Tanmay Gupta and
                  Kuo{-}Hao Zeng and
                  Jon Borchardt and
                  Dirk Groeneveld and
                  Jen Dumas and
                  Crystal Nam and
                  Sophie Lebrecht and
                  Caitlin Wittlif and
                  Carissa Schoenick and
                  Oscar Michel and
                  Ranjay Krishna and
                  Luca Weihs and
                  Noah A. Smith and
                  Hannaneh Hajishirzi and
                  Ross B. Girshick and
                  Ali Farhadi and
                  Aniruddha Kembhavi},
  title        = {Molmo and PixMo: Open Weights and Open Data for State-of-the-Art Multimodal
                  Models},
  journal      = {CoRR},
  volume       = {abs/2409.17146},
  year         = {2024},
  url          = {https://doi.org/10.48550/arXiv.2409.17146},
  doi          = {10.48550/ARXIV.2409.17146},
  eprinttype    = {arXiv},
  eprint       = {2409.17146},
  bibsource    = {dblp computer science bibliography, https://dblp.org}
}

@article{mammoth,
      title={MAmmoTH-VL: Eliciting Multimodal Reasoning with Instruction Tuning at Scale}, 
      author={Jarvis Guo and Tuney Zheng and Yuelin Bai and Bo Li and Yubo Wang and King Zhu and Yizhi Li and Graham Neubig and Wenhu Chen and Xiang Yue},
      year={2024},
      eprint={2412.05237},
      archivePrefix={arXiv},
      primaryClass={cs.CL},
      url={https://arxiv.org/abs/2412.05237}, 
}

@misc{sharegpt4o,
            author       = {Erfei Cui and Yinan He and Zheng Ma and Zhe Chen and Hao Tian and Weiyun Wang and Kunchang Li and Yi Wang and Wenhai Wang and Xizhou Zhu and Lewei Lu and Tong Lu and Yali Wang and Limin Wang and Yu Qiao and Jifeng Dai},
            title        = {ShareGPT-4o: Comprehensive Multimodal Annotations With GPT-4o},
            year         = {2024},
            url          = {https://sharegpt4o.github.io/}, 
          }

@inproceedings{youcook2,
  author       = {Luowei Zhou and
                  Chenliang Xu and
                  Jason J. Corso},
  editor       = {Sheila A. McIlraith and
                  Kilian Q. Weinberger},
  title        = {Towards Automatic Learning of Procedures From Web Instructional Videos},
  booktitle    = {Proceedings of the Thirty-Second {AAAI} Conference on Artificial Intelligence,
                  (AAAI-18), the 30th innovative Applications of Artificial Intelligence
                  (IAAI-18), and the 8th {AAAI} Symposium on Educational Advances in
                  Artificial Intelligence (EAAI-18), New Orleans, Louisiana, USA, February
                  2-7, 2018},
  pages        = {7590--7598},
  publisher    = {{AAAI} Press},
  year         = {2018},
  url          = {https://doi.org/10.1609/aaai.v32i1.12342},
  doi          = {10.1609/AAAI.V32I1.12342},
  bibsource    = {dblp computer science bibliography, https://dblp.org}
}

@inproceedings{crosstask,
  author       = {Dimitri Zhukov and
                  Jean{-}Baptiste Alayrac and
                  Ramazan Gokberk Cinbis and
                  David F. Fouhey and
                  Ivan Laptev and
                  Josef Sivic},
  title        = {Cross-Task Weakly Supervised Learning From Instructional Videos},
  booktitle    = {{IEEE} Conference on Computer Vision and Pattern Recognition, {CVPR}
                  2019, Long Beach, CA, USA, June 16-20, 2019},
  pages        = {3537--3545},
  publisher    = {Computer Vision Foundation / {IEEE}},
  year         = {2019},
  url          = {http://openaccess.thecvf.com/content\_CVPR\_2019/html/Zhukov\_Cross-Task\_Weakly\_Supervised\_Learning\_From\_Instructional\_Videos\_CVPR\_2019\_paper.html},
  doi          = {10.1109/CVPR.2019.00365},
  bibsource    = {dblp computer science bibliography, https://dblp.org}
}

@inproceedings{coin,
  author       = {Yansong Tang and
                  Dajun Ding and
                  Yongming Rao and
                  Yu Zheng and
                  Danyang Zhang and
                  Lili Zhao and
                  Jiwen Lu and
                  Jie Zhou},
  title        = {{COIN:} {A} Large-Scale Dataset for Comprehensive Instructional Video
                  Analysis},
  booktitle    = {{IEEE} Conference on Computer Vision and Pattern Recognition, {CVPR}
                  2019, Long Beach, CA, USA, June 16-20, 2019},
  pages        = {1207--1216},
  publisher    = {Computer Vision Foundation / {IEEE}},
  year         = {2019},
  url          = {http://openaccess.thecvf.com/content\_CVPR\_2019/html/Tang\_COIN\_A\_Large-Scale\_Dataset\_for\_Comprehensive\_Instructional\_Video\_Analysis\_CVPR\_2019\_paper.html},
  doi          = {10.1109/CVPR.2019.00130},
  bibsource    = {dblp computer science bibliography, https://dblp.org}
}

@article{idefics3,
  author       = {Hugo Lauren{\c{c}}on and
                  Andr{\'{e}}s Marafioti and
                  Victor Sanh and
                  L{\'{e}}o Tronchon},
  title        = {Building and better understanding vision-language models: insights
                  and future directions},
  journal      = {CoRR},
  volume       = {abs/2408.12637},
  year         = {2024},
  url          = {https://doi.org/10.48550/arXiv.2408.12637},
  doi          = {10.48550/ARXIV.2408.12637},
  eprinttype    = {arXiv},
  eprint       = {2408.12637},
  bibsource    = {dblp computer science bibliography, https://dblp.org}
}

@inproceedings{siglip,
  author       = {Xiaohua Zhai and
                  Basil Mustafa and
                  Alexander Kolesnikov and
                  Lucas Beyer},
  title        = {Sigmoid Loss for Language Image Pre-Training},
  booktitle    = {{IEEE/CVF} International Conference on Computer Vision, {ICCV} 2023,
                  Paris, France, October 1-6, 2023},
  pages        = {11941--11952},
  publisher    = {{IEEE}},
  year         = {2023},
  url          = {https://doi.org/10.1109/ICCV51070.2023.01100},
  doi          = {10.1109/ICCV51070.2023.01100},
  bibsource    = {dblp computer science bibliography, https://dblp.org}
}

@inproceedings{guo2025trace,
title={{TRACE}: Temporal Grounding Video {LLM}  via Causal Event Modeling},
author={Yongxin Guo and Jingyu Liu and Mingda Li and Qingbin Liu and Xi Chen and Xiaoying Tang},
booktitle={The Thirteenth International Conference on Learning Representations},
year={2025},
url={https://openreview.net/forum?id=14fFV0chUS}
}

@inproceedings{banerjee-lavie-2005-meteor,
    title = "{METEOR}: An Automatic Metric for {MT} Evaluation with Improved Correlation with Human Judgments",
    author = "Banerjee, Satanjeev  and
      Lavie, Alon",
    editor = "Goldstein, Jade  and
      Lavie, Alon  and
      Lin, Chin-Yew  and
      Voss, Clare",
    booktitle = "Proceedings of the {ACL} Workshop on Intrinsic and Extrinsic Evaluation Measures for Machine Translation and/or Summarization",
    month = jun,
    year = "2005",
    address = "Ann Arbor, Michigan",
    publisher = "Association for Computational Linguistics",
    url = "https://aclanthology.org/W05-0909/",
    pages = "65--72"
}

@inproceedings{vqav2,
  author       = {Yash Goyal and
                  Tejas Khot and
                  Douglas Summers{-}Stay and
                  Dhruv Batra and
                  Devi Parikh},
  title        = {Making the {V} in {VQA} Matter: Elevating the Role of Image Understanding
                  in Visual Question Answering},
  booktitle    = {2017 {IEEE} Conference on Computer Vision and Pattern Recognition,
                  {CVPR} 2017, Honolulu, HI, USA, July 21-26, 2017},
  pages        = {6325--6334},
  publisher    = {{IEEE} Computer Society},
  year         = {2017},
  url          = {https://doi.org/10.1109/CVPR.2017.670},
  doi          = {10.1109/CVPR.2017.670},
  bibsource    = {dblp computer science bibliography, https://dblp.org}
}

@article{fu2023mme,
  title={MME: A Comprehensive Evaluation Benchmark for Multimodal Large Language Models},
  author={Fu, Chaoyou and Chen, Peixian and Shen, Yunhang and Qin, Yulei and Zhang, Mengdan and Lin, Xu and Yang, Jinrui and Zheng, Xiawu and Li, Ke and Sun, Xing and others},
  journal={arXiv preprint arXiv:2306.13394},
  year={2023}
}

@inproceedings{gqa,
  author       = {Drew A. Hudson and
                  Christopher D. Manning},
  title        = {{GQA:} {A} New Dataset for Real-World Visual Reasoning and Compositional
                  Question Answering},
  booktitle    = {{IEEE} Conference on Computer Vision and Pattern Recognition, {CVPR}
                  2019, Long Beach, CA, USA, June 16-20, 2019},
  pages        = {6700--6709},
  publisher    = {Computer Vision Foundation / {IEEE}},
  year         = {2019},
  url          = {http://openaccess.thecvf.com/content\_CVPR\_2019/html/Hudson\_GQA\_A\_New\_Dataset\_for\_Real-World\_Visual\_Reasoning\_and\_Compositional\_CVPR\_2019\_paper.html},
  doi          = {10.1109/CVPR.2019.00686},
  bibsource    = {dblp computer science bibliography, https://dblp.org}
}

@misc{claude,
  author    = {Anthropic},
  title     = {Model Card Addendum: Claude 3.5 Haiku and Upgraded Claude 3.5 Sonnet},
  year      = {2024},
  month     = oct,
  url       = {https://assets.anthropic.com/m/1cd9d098ac3e6467/original/Claude-3-Model-Card-October-Addendum.pdf},
  note      = {Accessed: 2025-04-03}
}

@misc{MistralAITeam2024AuLarge,
  author       = {Mistral AI Team},
  title        = {Au Large},
  year         = {2024},
  month        = feb,
  url          = {https://mistral.ai/news/mistral-large},
  howpublished = {Mistral AI Blog Post},
  note         = {Accessed: 2025-04-03}
}

@incollection{mccloskey1989catastrophic,
  title={Catastrophic interference in connectionist networks: The sequential learning problem},
  author={McCloskey, Michael and Cohen, Neal J},
  booktitle={Psychology of learning and motivation},
  volume={24},
  pages={109--165},
  year={1989},
  publisher={Elsevier}
}

@article{Mathew2020DocVQAADA,
  title={DocVQA: A Dataset for VQA on Document Images},
  author={Minesh Mathew and Dimosthenis Karatzas and R. Manmatha and C. V. Jawahar},
  journal={2021 IEEE Winter Conference on Applications of Computer Vision (WACV)},
  year={2020},
  pages={2199-2208},
  url={https://api.semanticscholar.org/CorpusId:220280200}
}

@article{Lobry2020RSVQAVQA,
  author       = {Sylvain Lobry and
                  Diego Marcos and
                  Jesse Murray and
                  Devis Tuia},
  title        = {{RSVQA:} Visual Question Answering for Remote Sensing Data},
  journal      = {{IEEE} Trans. Geosci. Remote. Sens.},
  volume       = {58},
  number       = {12},
  pages        = {8555--8566},
  year         = {2020},
  url          = {https://doi.org/10.1109/TGRS.2020.2988782},
  doi          = {10.1109/TGRS.2020.2988782},
  bibsource    = {dblp computer science bibliography, https://dblp.org}
}

@article{Ren2020CGMVQAANA,
  title={CGMVQA: A New Classification and Generative Model for Medical Visual Question Answering},
  author={F. Ren and Yangyang Zhou},
  journal={IEEE Access},
  year={2020},
  volume={8},
  pages={50626-50636},
  url={https://api.semanticscholar.org/CorpusId:214691166}
}

@article{Tapaswi2015MovieQAUSA,
  title={MovieQA: Understanding Stories in Movies through Question-Answering},
  author={Makarand Tapaswi and Yukun Zhu and R. Stiefelhagen and A. Torralba and R. Urtasun and S. Fidler},
  journal={2016 IEEE Conference on Computer Vision and Pattern Recognition (CVPR)},
  year={2015},
  pages={4631-4640},
  url={http://ieeexplore.ieee.org/stamp/stamp.jsp?tp=\&arnumber=7780870}
}

@inproceedings{chen2024internvl,
  title={Internvl: Scaling up vision foundation models and aligning for generic visual-linguistic tasks},
  author={Chen, Zhe and Wu, Jiannan and Wang, Wenhai and Su, Weijie and Chen, Guo and Xing, Sen and Zhong, Muyan and Zhang, Qinglong and Zhu, Xizhou and Lu, Lewei and others},
  booktitle={Proceedings of the IEEE/CVF conference on computer vision and pattern recognition},
  pages={24185--24198},
  year={2024}
}

@misc{liu2024nvila,
      title={NVILA: Efficient Frontier Visual Language Models},
      author={Zhijian Liu and Ligeng Zhu and Baifeng Shi and Zhuoyang Zhang and Yuming Lou and Shang Yang and Haocheng Xi and Shiyi Cao and Yuxian Gu and Dacheng Li and Xiuyu Li and Yunhao Fang and Yukang Chen and Cheng-Yu Hsieh and De-An Huang and An-Chieh Cheng and Vishwesh Nath and Jinyi Hu and Sifei Liu and Ranjay Krishna and Daguang Xu and Xiaolong Wang and Pavlo Molchanov and Jan Kautz and Hongxu Yin and Song Han and Yao Lu},
      year={2024},
      eprint={2412.04468},
      archivePrefix={arXiv},
      primaryClass={cs.CV},
      url={https://arxiv.org/abs/2412.04468},
}

@article{foodllm,
  author       = {Yuehao Yin and
                  Huiyan Qi and
                  Bin Zhu and
                  Jingjing Chen and
                  Yu{-}Gang Jiang and
                  Chong{-}Wah Ngo},
  title        = {FoodLMM: {A} Versatile Food Assistant using Large Multi-modal Model},
  journal      = {CoRR},
  volume       = {abs/2312.14991},
  year         = {2023},
  url          = {https://doi.org/10.48550/arXiv.2312.14991},
  doi          = {10.48550/ARXIV.2312.14991},
  eprinttype    = {arXiv},
  eprint       = {2312.14991},
  bibsource    = {dblp computer science bibliography, https://dblp.org}
}

@article{biderman2024lora,
title={Lo{RA} Learns Less and Forgets Less},
author={Dan Biderman and Jacob Portes and Jose Javier Gonzalez Ortiz and Mansheej Paul and Philip Greengard and Connor Jennings and Daniel King and Sam Havens and Vitaliy Chiley and Jonathan Frankle and Cody Blakeney and John Patrick Cunningham},
journal={Transactions on Machine Learning Research},
issn={2835-8856},
year={2024},
url={https://openreview.net/forum?id=aloEru2qCG},
note={Featured Certification}
}

@article{llavaonevision,
  title={Llava-onevision: Easy visual task transfer},
  author={Li, Bo and Zhang, Yuanhan and Guo, Dong and Zhang, Renrui and Li, Feng and Zhang, Hao and Zhang, Kaichen and Zhang, Peiyuan and Li, Yanwei and Liu, Ziwei and others},
  journal={arXiv preprint arXiv:2408.03326},
  year={2024}
}

@misc{bai2025qwen25vltechnicalreport,
      title={Qwen2.5-VL Technical Report}, 
      author={Shuai Bai and Keqin Chen and Xuejing Liu and Jialin Wang and Wenbin Ge and Sibo Song and Kai Dang and Peng Wang and Shijie Wang and Jun Tang and Humen Zhong and Yuanzhi Zhu and Mingkun Yang and Zhaohai Li and Jianqiang Wan and Pengfei Wang and Wei Ding and Zheren Fu and Yiheng Xu and Jiabo Ye and Xi Zhang and Tianbao Xie and Zesen Cheng and Hang Zhang and Zhibo Yang and Haiyang Xu and Junyang Lin},
      year={2025},
      eprint={2502.13923},
      archivePrefix={arXiv},
      primaryClass={cs.CV},
      url={https://arxiv.org/abs/2502.13923}, 
}

@inproceedings{actionabot,
author = {Zhu, Qingxiaoyang and Lee, Yi-Chieh and Wang, Hao-Chuan},
title = {Action-a-Bot: Exploring Human-Chatbot Conversations for Actionable Instruction Giving and Following},
year = {2022},
isbn = {9781450391900},
publisher = {Association for Computing Machinery},
address = {New York, NY, USA},
url = {https://doi.org/10.1145/3500868.3559476},
doi = {10.1145/3500868.3559476},
booktitle = {Companion Publication of the 2022 Conference on Computer Supported Cooperative Work and Social Computing},
pages = {145–149},
numpages = {5},
location = {Virtual Event, Taiwan},
series = {CSCW'22 Companion}
}

@inproceedings{Zhu2025ActionaBotSM,
  title={ActionaBot: Structuring Metacognitive Conversations towards In-Situ Awareness in How-To Instruction Following},
  author={Qingxiaoyang Zhu and Yi-Chieh Lee and Hao-Chuan Wang},
  booktitle={International Conference on Conversational User Interfaces},
  year={2025},
  url={https://api.semanticscholar.org/CorpusID:279962728}
}

@misc{bai2025qwen3vltechnicalreport,
      title={Qwen3-VL Technical Report}, 
      author={Shuai Bai and Yuxuan Cai and Ruizhe Chen and Keqin Chen and Xionghui Chen and Zesen Cheng and Lianghao Deng and Wei Ding and Chang Gao and Chunjiang Ge and Wenbin Ge and Zhifang Guo and Qidong Huang and Jie Huang and Fei Huang and Binyuan Hui and Shutong Jiang and Zhaohai Li and Mingsheng Li and Mei Li and Kaixin Li and Zicheng Lin and Junyang Lin and Xuejing Liu and Jiawei Liu and Chenglong Liu and Yang Liu and Dayiheng Liu and Shixuan Liu and Dunjie Lu and Ruilin Luo and Chenxu Lv and Rui Men and Lingchen Meng and Xuancheng Ren and Xingzhang Ren and Sibo Song and Yuchong Sun and Jun Tang and Jianhong Tu and Jianqiang Wan and Peng Wang and Pengfei Wang and Qiuyue Wang and Yuxuan Wang and Tianbao Xie and Yiheng Xu and Haiyang Xu and Jin Xu and Zhibo Yang and Mingkun Yang and Jianxin Yang and An Yang and Bowen Yu and Fei Zhang and Hang Zhang and Xi Zhang and Bo Zheng and Humen Zhong and Jingren Zhou and Fan Zhou and Jing Zhou and Yuanzhi Zhu and Ke Zhu},
      year={2025},
      eprint={2511.21631},
      archivePrefix={arXiv},
      primaryClass={cs.CV},
      url={https://arxiv.org/abs/2511.21631}, 
}

@article{wang2025internvl3,
  title={Internvl3.5: Advancing open-source multimodal models in versatility, reasoning, and efficiency},
  author={Wang, Weiyun and Gao, Zhangwei and Gu, Lixin and Pu, Hengjun and Cui, Long and Wei, Xingguang and Liu, Zhaoyang and Jing, Linglin and Ye, Shenglong and Shao, Jie and others},
  journal={arXiv preprint arXiv:2508.18265},
  year={2025}
}

\clearpage

\appendix

\section{Training Dataset Details}
\label{app:app_datasets}

\begin{table*}[t]
  \centering
  \small
  \caption{Overview of the multistage training process for \modelname. Status indicators: \fireemoji = Trainable, \iceemoji = Frozen.}
  \label{tab:training_stages}
  \begin{tabularx}{0.95\textwidth}{l c c c X}
    \toprule
    \multirow{2}{*}{\textbf{Stage}} & \textbf{Specialized} & \multirow{2}{*}{\textbf{LLM}} & \multirow{2}{*}{\textbf{VE}} & \multirow{2}{*}{\textbf{Datasets}} \\
      & \textbf{Layers} &   &   &   \\
    \midrule
    Initialization & \fireemoji & \iceemoji & \iceemoji & Laion400M~\cite{laion400m} \\ \addlinespace
    Visual Instruction Tuning & \fireemoji & \fireemoji & \fireemoji & ShareGPT4o~\cite{sharegpt4o}, MAmmoTH-VL~\cite{mammoth}, VQAv2~\cite{vqav2}, GQA~\cite{gqa}, ShareGPT4v~\cite{sharegpt4v}, PixMo-Cap~\cite{pixmo}, LLaVA-Pretrain~\cite{llava} \\ \addlinespace
    Domain-specific Tuning & \fireemoji & \fireemoji & \iceemoji & FoodDialogues~\cite{foodllm}, FoodReasonSeg~\cite{foodllm}, Youcook2~\cite{youcook2}, COIN~\cite{coin}, Crosstask~\cite{crosstask} \\ \addlinespace
    Task-specific Training & \fireemoji & \fireemoji & \iceemoji & \datasetname{} (ours) \\
    \bottomrule
  \end{tabularx}
\end{table*}

\begin{table}[]
  \centering
  \caption{Overview of the datasets used to train \modelname{}.}
  \label{tab:app_datasets}
  \small
  \begin{tabular}{@{}clcc@{}}
  \toprule
  \textbf{Stage}              & \textbf{Dataset}        & \textbf{Type} & \textbf{Samples \#}  \\ \midrule
  \textbf{1}                  & LAION400M      & Cap + Ret & 10M+10M        \\ \midrule
  \multirow{7}{*}{\textbf{2}} & ShareGPT4o     & Cap & $\approx$57k           \\
                     & MAmmoTH-VL        & Reasoning & 1M           \\
                     & VQAv2          & pVQA & $\approx$444k           \\
                     & GQA            & pVQA & 943k           \\
                     & ShareGPT4v     & Ret & $\approx$1.25M          \\
                     & Pixmo-cap      & Ret & $\approx$698k         \\
                     & LLaVA-Pretrain & Ret & $\approx$558k           \\ \midrule
  \multirow{5}{*}{\textbf{3}} & FoodDialogues  & Domain & $\approx$26k           \\
                     & FoodReasonSeg  & Domain & $\approx$14k            \\
                     & YouCook2       & Cap+Ret & $\approx$10k+$\approx$10k            \\
                     & Crosstask      & Cap+Ret & $\approx$17k+$\approx$16k           \\
                     & COIN           & Cap+Ret & $\approx$43k+$\approx$39k            \\ \midrule
  \textbf{4}         & \datasetname{}  & All & $\approx$110k           \\ \bottomrule
  \end{tabular}
\end{table}

In this section, we provide additional details on the datasets used to train \modelname{}. We include the specific number of samples used in each stage, as shown in Table~\ref{tab:app_datasets}.

\subsection{Multitask Trainer}

As there are 2 losses being optimized in our model, passing through task-specific components, we need to employ a dual dataloader trainer that allows us to train on both tasks at the same time without needing to format our batches in a multitask format, something that is not doable for pVQA, Visual Instruction Tuning, and CMVR. This is achieved by alternating the dataloaders when sampling batches, allowing for an even distribution of training batches.

\subsection{Dataset Transformations}

As we target instructional plans in a conversational manner, we needed to adapt and transform some of the datasets to train our model. In this section, we detail the transformations applied to each dataset:

\textbf{LAION400M} - For LAION400M we use the image-caption pairs to train the model in a Image-to-Text task and Text-to-Image retrieval.

\textbf{ShareGPT4o} - This dataset is used as is, with no alterations.

\textbf{MAmmoTH-VL} - This dataset is used as is, with no alterations. We only consider the first 1M multimodal samples.

\textbf{VQAv2 \& GQA} - These datasets are framed as an instruction following task using templates to convert the question into an instruction, the answer is kept as is.

\textbf{ShareGPT4v, LLaVA-Pretrain, Pixmo-cap} - These datasets are used as is, with no alterations. Pixmo is framed as a user request for a detailed image caption.

\textbf{FoodDialogues} - This dataset includes special tags, such as \texttt{[CARB]}, to indicate nutritional info estimation in system answers, we remove these tags and preserve the rest of the text. 

\textbf{FoodReasonSeg} - This dataset includes requests for segmentation mask along with ingredient identification, we focus on the latter and remove the segmentation mask requests. We also rewrite the user utterances using a Llama 3 model, to improve naturalness and fluency.

\textbf{YouCook2, Crosstask, COIN} - These datasets are made up of instructional videos with annotated video segments. To adapt them to out needs, for each annotated video segment, we extract 4 frames and train the model to predict the annotated text. For retrieval, we instead provide the model with the task name and annotated video segment text and train the model to retrieve the start and end frames of the video segment. In this case we adopt the strategy of picking $N$ frames from the start and end, and the model is trained to approximate the average of the embeddings of these frames. This allows us to train the model to retrieve a single frame, while also allowing for some flexibility in the candidate video moment limits.

\section{Training Details}
\label{app:train_details}

\begin{table}[]
  \caption{Shared hyperparameters used to train all models. For Stage 1, LLM-specific hyperparameters only affect the \texttt{[RET]} tokens input and output embeddings, along with the language modeling head.}
  \label{tab:app_shared_hyperparameters}
  \centering
  \resizebox{\linewidth}{!}{%
  \begin{tabular}{l|cccc}
  \toprule
  \textbf{Stage}              & 1       & 2          & 3        & 4 \\ \midrule
  Batch Size                  & 512     & 64         & 128       & 32        \\
  RET Batch Size              & 512     & 96         & 192       & ---        \\
  Acc. Steps                  & 4/4/8   & 4/8/8      & 4/8/8     & 4/8/16     \\
  Train Steps                 & 19500   & 15000      & 866       & 3252        \\
  Parallel                    & DDP     & DDP        & DDP       & DDP       \\
  GPU \#                      &   4     & 8          & 4         & 1          \\
  Model DType                 & BF16  & BF16      & BF16    & BF16 \\ \midrule
  LLM LR                      & $1*10^{-3}$ & $5*10^{-5}$     & $2*10^{-6}$    & $5*10^{-5}$    \\ 
  VE LR                       & ---         & $5*10^{-5}$     & ---            & ---    \\
  CAP LR                      & $5*10^{-4}$ & $5*10^{-4}$     & $5*10^{-4}$    & $2*10^{-4}$    \\
  RET LR                      & $1*10^{-3}$ & $1*10^{-3}$     & $1*10^{-3}$    & $1*10^{-3}$    \\ \midrule
  Scheduler                   & Constant& Constant    & Constant  & Constant           \\
  Optimizer                   & AdamW   & AdamW       & AdamW     & AdamW     \\
  LLM Decay                   & 0.03    & 0.01        & 0.0       & 0.0       \\
  VE Decay                    & ---     & 0.0         & ---       & ---       \\
  CAP Decay                   & 0.03    & 0.03        & 0.01      & 0.01      \\
  RET Decay                   & 0.03    & 0.03        & 0.01      & 0.01      \\ \midrule
  \multicolumn{5}{c}{\textbf{LoRA (Only for 8B)}} \\ \midrule
  LoRa DType                  & ---     & FP32        & FP32      & FP32      \\
  LoRa Rank                   & ---     & 32          & 32        & 32        \\
  LoRa $\alpha$               & ---     & 128         & 128       & 128       \\
  LoRa Dropout                & ---     & 0.03        & 0.03      & 0.03      \\ \bottomrule
  \end{tabular}}
\end{table}

In this section, we provide additional details on the training setup and hyperparameters used to train \modelname{}. The hyperparameters are shared across all models and are shown in Table~\ref{tab:app_shared_hyperparameters}, for each training stage. The training was performed using NVIDIA A100 GPUs with 80GB of memory each, connected with SXM4. 

On stage 2 and 3 we leverage our dual dataloader training approach to utilize larger batch sizes for retrieval, this was possible as the retrieval samples are smaller than the captioning samples. This approach benefits retrieval training as we utilize in-batch samples as negatives for the contrastive loss calculation.

For stage 4 we use a single dataloader with a batch size of 32, and the different mode samples are uniformly sampled from the dataset, with this sampling logic being handled by the sampler instead.

In Table~\ref{tab:app_shared_hyperparameters}, CAP refers to the connector between the Vision Encoder and the LLM, RET refers to the retrieval-specific layers used in the model ($W_{start}, W_{end}, W_{ret}$).

\section{LM Backbone Size Variance}
\label{app:lm_size}

\begin{table*}[!t]
\centering
\caption{Ablation of LM backbone sizes on model performance across all considered tasks.}
\label{tab:lm_size_ablations}
\begin{tabular}{@{}lcc|ccccc|ccccc@{}}
\toprule
 & \multicolumn{4}{c}{\textbf{Plan Reasoning}} & & \multicolumn{4}{c}{\textbf{Plan-based Retrieval}} & & \multicolumn{2}{c}{\textbf{General}} \\
 \cmidrule{2-5} \cmidrule{7-10} \cmidrule{12-13}
 & \multicolumn{2}{c|}{\textbf{PGAG}} & \multicolumn{2}{c}{\textbf{pVQA}} & & \multicolumn{2}{c|}{\textbf{VSG}} &\multicolumn{2}{c}{\textbf{CVMR R@1}} & & \multicolumn{2}{c}{\textbf{MME}}\\
 & \textit{R-L} & \textit{METEOR} & \textit{R-L} & \textit{Acc.} & & \textit{R-L} & \textit{EM} & \textit{m=0.5} & \textit{m=0.7} & & \textit{Cog.} & \textit{Per.} \\ \midrule
 \modelname{} 1B & 75.44 & 76.67 & 29.35 & 58.69 & & 51.53 &  & 26.22 & 13.92 & & 203 & 632 \\
 \modelname{} 3B & \textbf{76.51} & \textbf{77.72} & 30.16 & 71.59 & & 52.15 & 53.68 & 23.33 & \textbf{15.91} & & 214 & \textbf{808}  \\
 \modelname{} 8B & 75.30 & 76.67 & \textbf{33.65} & \textbf{94.02} & & \textbf{55.66} & \textbf{57.37} & \textbf{30.74} & 15.37 & & \textbf{229} & 717  \\ 
 \bottomrule
\end{tabular}
\end{table*}

To measure how model performance varies across LM backbone sizes, we ran some experiments with other Llama 3 model sizes. Specifically, we tested with Llama 3 1B and 3B, the results for all tasks are shown in Table~\ref{tab:lm_size_ablations}.

Here we see that, with some exceptions, the smaller models have very strong performances, considering their reduced size, whilst also being clear the benefit of increase LM backbone size. For \textbf{CVMR} \modelname{} 1B outperforms 3B, we hypothesize that it is caused by 3B training with a smaller batch size, which slows down convergence as InfoNCE loss uses in-batch negatives, meaning that smaller batches can slow down convergence. Focusing on \textbf{pVQA} the benefit of higher parameter count is very accentuated with significant improvements as we scale up the LM. As this is a task that requires the models to encode external knowledge acquired during training thus smaller models have more limited external knowledge remembrance capacity. Whereas for \textbf{PGAG}, where most requests are context-dependent, the models all have impressively similar performance. \textbf{MME} emerges as an outlier, having similar performance to the 3B variant, despite the increased LLM size. We hypothesize that this is caused by training the 8B LM backbone using LoRA~\cite{hu2022lora}, due to memory constraints, from stage 2 onward, which means that only a small number of parameters are being updated, thus suffering from decreased learning capacity, as studied in~\citet{biderman2024lora}.

\section{CVMR Evaluation}
\label{app:cvmr_threshold}

\begin{algorithm}[t]
\caption{Candidate video moment extraction in the \textit{``adjusted"} setting.}
\label{alg:adjusted_extraction}
\small
\SetAlgoLined
\DontPrintSemicolon

\SetKwInOut{Input}{Input}
\SetKwInOut{Output}{Output}

\Input{Start representation $r_{start}$,\\End representation $r_{end}$, \\ Encoded Frames $\mathcal{V} = \{v_1, v_2, \dots, v_T\}$, \\ Similarity threshold $\tau$}
\Output{Temporal segment indices $(t_{start}, t_{end})$}

\BlankLine
$idx_{s} \leftarrow \operatorname*{arg\,max}_{i \in \{1, \dots, T\}} \operatorname{Sim}(v_i, r_{start})$\;
$idx_{e} \leftarrow \operatorname*{arg\,max}_{i \in \{1, \dots, T\}} \operatorname{Sim}(v_i, r_{end})$\;

\BlankLine
$t_{start} \leftarrow idx_{s}$\;
\For{$i \leftarrow idx_{s} \text{ \textbf{down to} } 1$}{
    \If{$\operatorname{Sim}(v_i, r_{start}) < \tau$}{
        \textbf{break} \tcp*{Stop if similarity drops}
    }
    $t_{start} \leftarrow i$\;
}

\BlankLine
$t_{end} \leftarrow idx_{e}$\;
\For{$i \leftarrow idx_{e} \text{ \textbf{to} } T$}{
    \If{$\operatorname{Sim}(v_i, r_{end}) < \tau$}{
        \textbf{break} \tcp*{Stop if similarity drops}
    }
    $t_{end} \leftarrow i$\;
}

\BlankLine
\Return{$(t_{start}, t_{end})$}\;

\end{algorithm}

Although the task of CVMR is formulated by retrieving the start and end frame of a relevant video moment, realizing the lack of flexibility of such an approach, specifically in a setting where adjacent images are largely similar, we train our model to instead produce the representation of the averaged N frames around the target frame. Based on this, it is fair to say that our model is trained to predict a start and end region representation rather than only retrieving a individual start and end frames. 

To account for this, during evaluation, we evaluate our models and MM-PlanLLM in what we call an \textit{``adjusted"} candidate extraction setting where the limits of the candidate video moment are determined by the point/frame where the frame's similarity to the produced representation drops below a given threshold. Specifically, in the \textit{``adjusted"} approach, the extraction of candidate video moments follows the logic of Algorithm~\ref{alg:adjusted_extraction}.

This leverages the observations of~\citet{mmplanllm} that the model's retrieval features show a decrease in similarity to video frames belonging to steps further away from the target step, suggesting that the model is able to effectively identify the target plan step and produce an adequate video region representation.
This threshold is determined using the validation set, as seen in Figure~\ref{fig:cvmr_threshold_main}, where we report R@1 values for mIoU = 0.5 and 0.7, with 20 different threshold values. Here we observe a constant CVMR performance improvement until threshold = 0.35 followed by a sharp decline when the threshold > 0.35, showing that the model has clearly learned region-level features. Interestingly, when mIoU=0.7 instead of a peak at 0.35 we see a plateau from 0.2 to 0.35, which reveals that the model is able to identify the moment’s core content at around 0.35 similarity with its outer limits having between 0.2 and 0.35 similarity with the retrieval features. The existence of a plateau reinforces that the model is capable of isolating the relevant video moment. Based on this, we set the threshold to 0.35, where the model is at or close to its performance peak.

\begin{figure*}[t]

    \caption{Six examples of how the similarity of the start and end retrieval representations with the video frames varies throughout the video.}
    \label{fig:six_cmvr_plots}
    \centering

    \begin{subfigure}[b]{0.495\textwidth}
        \centering
        \includegraphics[width=1.05\linewidth]{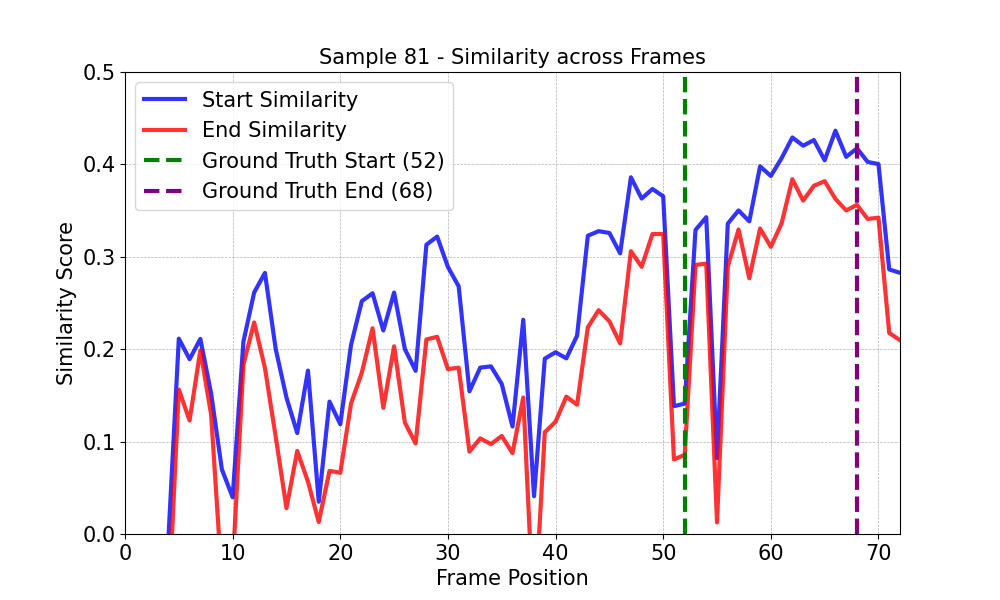}
        \caption{}
        \label{fig:1a}
    \end{subfigure}
    \hfill
    \begin{subfigure}[b]{0.495\textwidth}
        \centering
        \includegraphics[width=1.05\linewidth]{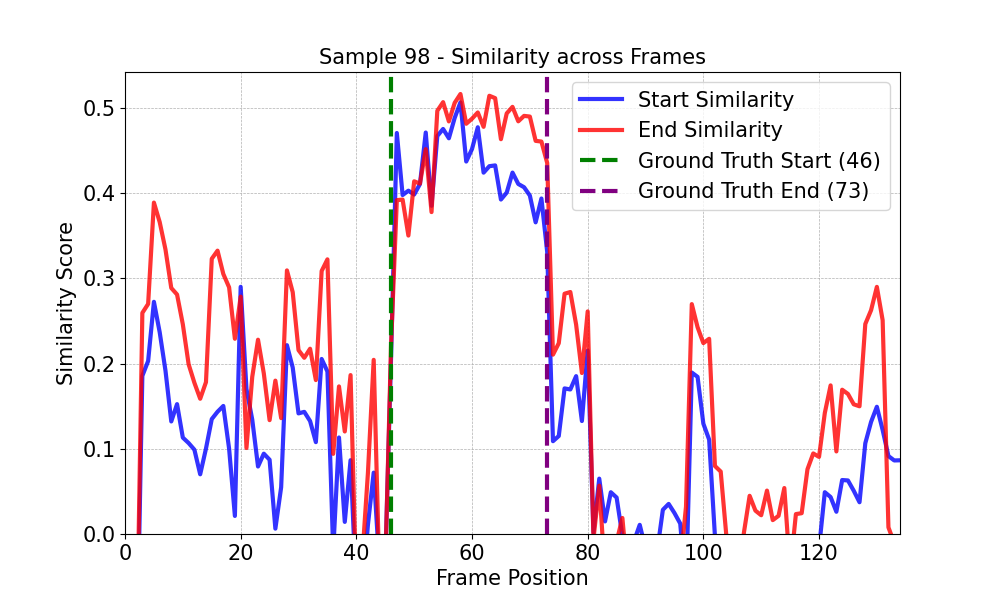}
        \caption{}
        \label{fig:1b}
    \end{subfigure}

    \vspace{1.5em}

    \begin{subfigure}[b]{0.495\textwidth}
        \centering
        \includegraphics[width=1.05\linewidth]{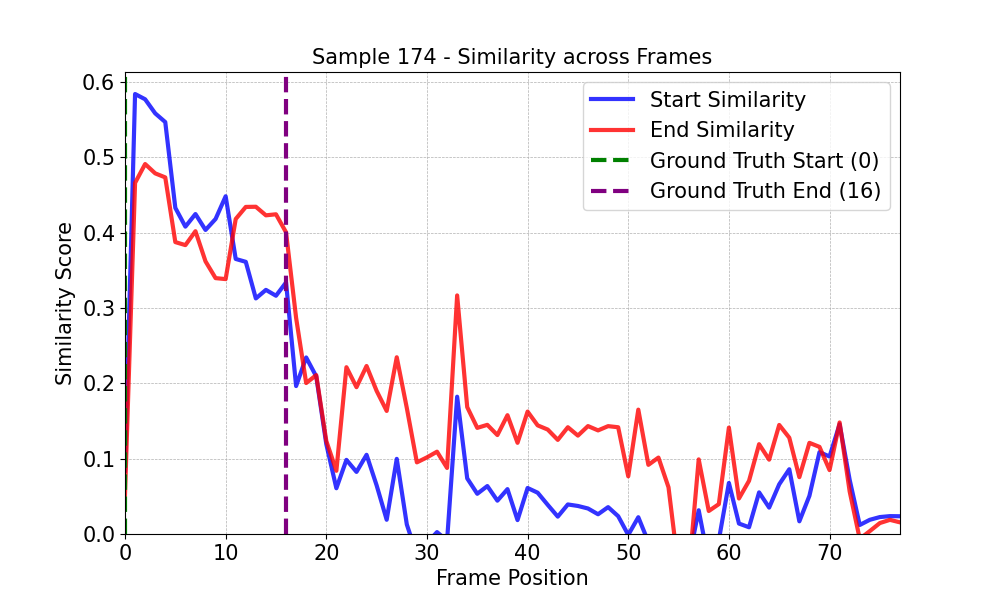}
        \caption{}
        \label{fig:1c}
    \end{subfigure}
    \hfill
    \begin{subfigure}[b]{0.495\textwidth}
        \centering
        \includegraphics[width=1.05\linewidth]{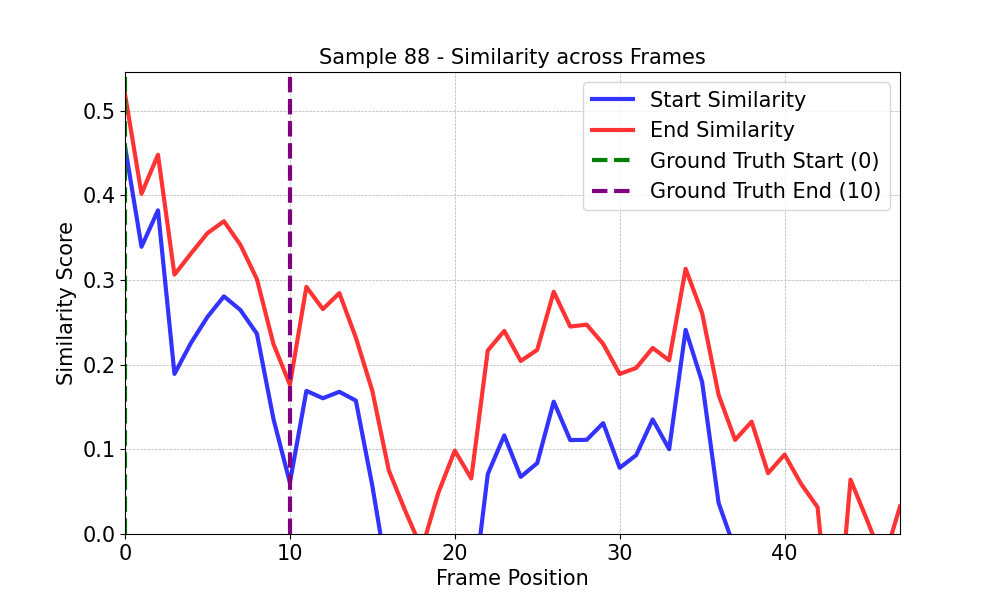}
        \caption{}
        \label{fig:1d}
    \end{subfigure}

    \vspace{1.5em}

    \begin{subfigure}[b]{0.495\textwidth}
        \centering
        \includegraphics[width=1.05\linewidth]{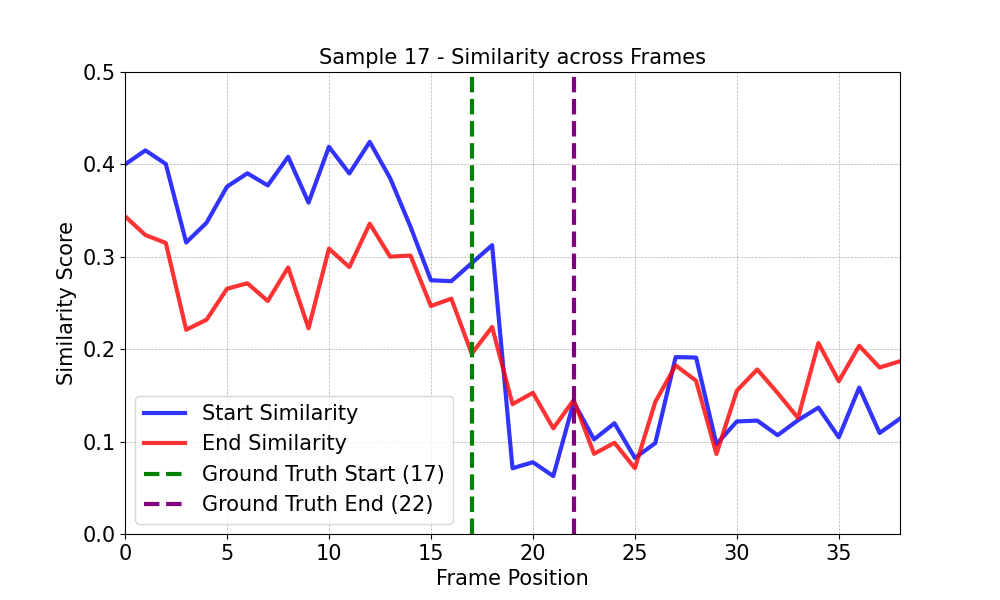}
        \caption{}
        \label{fig:1e}
    \end{subfigure}
    \hfill
    \begin{subfigure}[b]{0.495\textwidth}
        \centering
        \includegraphics[width=1.05\linewidth]{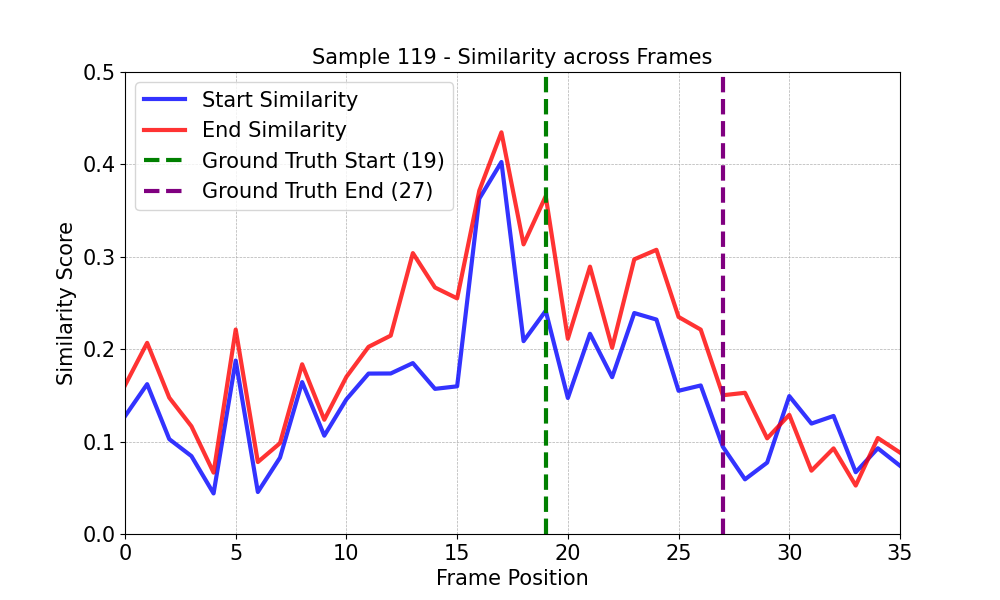}
        \caption{}
        \label{fig:1f}
    \end{subfigure}
\end{figure*}

To illustrate how similarity with the start and end retrieval representations varies across the video frames we included six plots in Figure~\ref{fig:six_cmvr_plots}. In these plots, the x axis is the frame position, and the y axis is the similarity score with the start (blue line) and end (red line) retrieval representations. Additionally, we plot two vertical lines that mark the ground truth video moment limits.

These plots provide an insight into how accurately the produced retrieval representations line up against the instructional video. In the first few examples we see clear cases where the produced representations have their similarity maximized in the target video moment frames, meaning that they correctly identified the video moment. Particularly in the ~\ref{fig:1b} plot where there's a clear peak and plateau spanning the target moment almost exactly. The plots also show how the model is able to produce retrieval representations that "track" video progress with similarity monotonically increasing and decreasing when getting closer and further from the target moment, respectively, as seen in plots~\ref{fig:1a} and ~\ref{fig:1c}. This same pattern is visible in~\ref{fig:1d} however here the similarity seems to start dropping slightly earlier than the end of the target video moment. The plots~\ref{fig:1e} and ~\ref{fig:1f} also showcase this pattern, however here the representations maximize similarity on the moment immediately before the target moment, leading to incorrect candidate extraction. Overall, these plots show that the model is capable of producing very accurate retrieval representations that correctly identify the position of the target video moment in most cases. We also see that there are similarity "valleys" separating the moments, showing that the model is able to model a clear moment-to-moment barrier.

\section{Text-only Ablation}
\label{app:text_only}

To measure the impact of multimodal training on the final text-only, PGAG, performance we also trained a llama directly on text-only in-domain data, meaning the model training solely on stage 4 and only on samples that do not involve multimodal inputs or outputs. We report these results in Table~\ref{tab:text_only_ablation}.

\begin{table}[!t]
\centering
  \caption{Results comparing the full model training pipeline against a text-only training run. \underline{Underlined} scores are the second best scores.}
  \label{tab:text_only_ablation}
  \small
  \begin{tabular}{lccc}
  \toprule
  & \multicolumn{3}{c}{\textbf{PGAG}} \\
  \textbf{Model} & \textit{ROUGE-L} & \textit{METEOR} & \textit{BERTScore} \\ \midrule
  Idefics2             & 37.16 & 45.09 & 67.55  \\
  LLaVA-1.5            & 42.47 & 40.84 & 70.55 \\
  Qwen 2.5 VL          & 31.61 & 41.51 & 63.91 \\
  LLaVA-OV             & 41.18 & 40.85 & 70.70 \\
  InternVL 3.5         & 41.62 & 46.20 & 69.22 \\
  Qwen 3 VL            & 44.71 & 46.31 & 71.94 \\
  MM-PlanLLM           & \underline{58.85} & \underline{59.34} & \underline{80.03} \\
  \addlinespace[0.5ex]
  \cdashline{1-4}
  \addlinespace[0.5ex]
  \modelname{}         & \textbf{75.30}   & \textbf{76.67}     & \textbf{88.72} \\ 
  \modelname{}-Text    & 75.09 & 76.47 & 88.68 \\ 
  \bottomrule
  \end{tabular}
\end{table}

From these results, we can see a slight negative performance delta against \modelname{}. This is likely due to the lack of the added multimodal training data that still exposes the model to the procedural plan guidance conversational patterns. This combined with the sacrifice of all multimodal capabilities, makes it far less capable for full procedural plan guidance.

\section{pVQA Generation Details}
\label{app:vqa_gen_details}
Of the two crucial extensions that we make to TastyVidDial is the addition of pVQA user requests. To achieve this, we used Claude 3.5 Sonnetv2 to generate user questions relevant to a provided image, the ongoing dialogue, and the instructional plan being executed. To determine whether a pVQA turn would be added we set a 30\% chance of that after any time the user moved forward in the task (using the NextStepIntent annotation provided in TastyVidDial). Table~\ref{tab:app_vqa_prompt} shows the prompt used to generate the visual question-answer pairs, and Table~\ref{tab:app_vqa_example} shows an example from \datasetname{}'s test set.

\section{pVQA Data Quality Annotation}
\label{app:app_data_anno}

To ascertain the quality of the generated turns we conducted a data quality annotation study followed by some error analysis study.

\subsection{Quality Annotation}

\begin{table}[!tpb]
\centering
\caption{Results of the data quality annotation conducted on 250 pVQA turns from the test set of \datasetname{}.}
\label{tab:data_quality_res}
\begin{tabular}{@{}lll@{}}
\toprule
\textbf{Plausibility} & \textbf{Relevance} & \textbf{Accuracy} \\ \midrule
2.22 & 2.67 & 2.75 \\ \bottomrule
\end{tabular}
\end{table}

As LLMs were used to generate the pVQA turns in \datasetname{} we find it important to conduct a data quality annotation study to ascertain the quality of the added turns on several metrics.
To this end, we conduct an annotation study with five volunteer annotators and tasked them with annotating the following 3 criteria:

\begin{itemize}
    \item \textbf{Plausibility} - Here we asked annotators to rate how plausible each turn is when considering the provided dialogue context in a real scenario. With this criteria, we aimed to measure how realistic the generated turns are and how well they pass for real user inputs.
    \item \textbf{Relevance} - For this criteria, we wanted to assess how relevant the generate user visual question was to both the selected image, and the instructional plan being executed. This is critical to understand if the generated turns are closely related to all of the surrounding context. 
    \item \textbf{Accuracy} - It was also important to ensure that the generated turns were made up of questions that were adequately answered and did not contain inaccurate answers instead. In this criteria, we asked annotators to, considering the dialogue, instructional plan, and their own knowledge, rate if the provided answers were accurate to the generated user questions.
\end{itemize}

The selected annotators were volunteers all with higher education in the computer science field, all fluent in English, and familiar with data annotation tasks and user studies.
In total 250 pVQA samples were randomly selected from \datasetname{}'s test set. Annotators were shown the instructional plan, provided user image, generated user question, and generated system response. 
The specific instructions provided to the volunteers are shown in Table~\ref{tab:anno_instructions}.

\begin{table}[h]
\centering
\caption{Annotation instructions provided to annotate in-domain pVQA turns.}
\label{tab:anno_instructions}
{\texttt{%
\begin{tabular}{p{0.98\linewidth}}
\toprule
Consider a dataset of dialogues between users and a system designed to assist with manual tasks (recipes or DIY projects). The objective of these annotations is to quantify the quality of automatically generated VQA (Visual Question Answering) turns within that dataset.\\
\\
Each line contains an example of a question-answer pair, along with the image that originates the question and the task to be performed.\\
For each line, on a scale of 1 to 3 (from worst to best), annotate the following 3 criteria:\\
\\
\textbf{Relevance}: The relevance of the question concerning the image and the task in question.\\
1 - The question has nothing to do with either the image or the task.\\
3 - The question is relevant to both.\\
\\
\textbf{Plausibility}: How plausible the question is, considering a person trying to perform the task.\\
1 - A question that no one would ask in the context of that task.\\
3 - A completely plausible question to be asked during the execution of the task.\\
\\
\textbf{Accuracy}: Based on the task and your own knowledge, how correct the given answer to the question is.\\
1 - An incorrect answer.\\
3 - A correct answer.\\
\\
Consider each criterion separately. A question may not be at all relevant or plausible (in the context of the task) but still have a correct answer.\\
\bottomrule
\end{tabular}}}
\end{table}

The results, shown in Table~\ref{tab:data_quality_res}, highlight the quality of the generated pVQA turns with very high scores for both Relevance and Accuracy, meaning that the turns were closely related to the context they were inserted in and provided reliable answers. Plausibility scores were also very positive asserting that the generated user questions often fell into the scope of possible questions a user might ask in a realistic setting, a key factor to ensure high data quality.

\subsection{Error Analysis}

To understand the possible failure patterns presented in the generated data, we manually examined the worst 50 (as measured by the average score of the 3 criteria annotated) turns from the ones in the data quality annotation and noted the most common patterns: \textbf{1.} The generated question phrases the image as not being the user's (20 occurrences) - In these cases, the question is framed as if the user is seeing the image somewhere, not that it is their image. \textbf{2.} Image contains text on screen (12 occurrences) – As images are extracted from instructional videos, occasionally they may contain some text on screen that renders their usage implausible for a user-uploaded image. \textbf{3.} Image-text mismatch (12 occurrences) – In some cases, the question mentions aspects (objects, tools, ingredients, etc.) that are not visible in the image.

To mitigate the above patterns, several approaches can be adopted:

\begin{itemize}
    \item The first failure pattern can be tackled by identifying common wording patterns in these cases and automatically rejecting samples with these patterns. Alternatively, a second LLM could be deployed with the sole task of annotating each sample on several criteria to automatically isolate low quality samples.
    \item The second pattern could be addressed by using an OCR model to try and extract text from the image. If text is successfully and confidently extracted, the image would be deemed unfit for usage.
    \item The third pattern poses a more challenging obstacle to address with no clear solution without relying on LVLMs, as it requires the extraction of entities in text and image and cross-referencing them to identify possible mismatches.
\end{itemize}

However, extrapolating this to the entire dataset, we highlight that this occurs in a small part of the dataset of \textbf{less than 10\% of the samples} having at least one of these.

\section{pVQA Evaluation Details}
\label{app:vqa_eval}

For the evaluation of pVQA turns on the \datasetname{} dataset, we utilize accuracy as measured by three LLMs. For this evaluation, we consider three large-scale LVLMs to maximize the quality of the annotations. Specifically, we utilize Claude 3.5 Sonnet v2, LLaMa Vision Instruct 90B, and Mistral Large.

For each model, we provided it with the relevant image, the previous 4 dialogue turns, and the instructional plan being followed. The prompt structure is present in Table~\ref{tab:app_vqa_eval_prompt}. The majority vote of the three responses is then used to provide the final accuracy score.

\begin{table*}[!tbp]
\caption{pVQA test set examples}
\label{tab:app_vqa_example}
\centering
\begin{tabular}{l|l|p{0.60\textwidth}}
\toprule
\textbf{User Image} & & \\ \midrule
\multirow{2}{*}{\raisebox{-\height}{\includegraphics[width=0.2\textwidth]{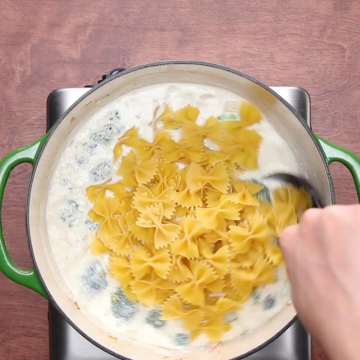}}} & \textbf{Question:} & I'm adding the pasta to the boiling milk mixture now - is it normal that the liquid looks this white and creamy already, or did I add too much milk? \\[4ex]
& \textbf{GT Answer:} & The liquid consistency shown in the image looks perfect! The white, creamy appearance is exactly what you want at this stage. The milk should be bubbling and have this rich, creamy texture before adding the farfalle (bow-tie) pasta. Just remember to keep stirring frequently while the pasta cooks for about 10 minutes at medium heat to prevent it from sticking to the bottom of the pot and to ensure even cooking. \\[4ex] \midrule

\multirow{2}{*}{\raisebox{-0.85\height}{\includegraphics[width=0.2\textwidth]{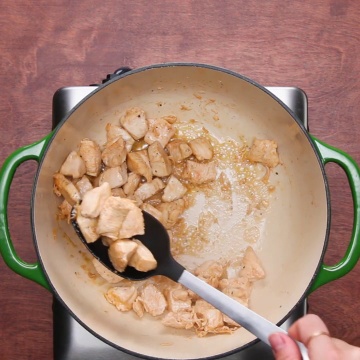}}} & \textbf{Question:} & Based on the color of the chicken pieces in the pan, are they browned and cooked through enough to be removed from the pan? \\[4ex]
& \textbf{GT Answer:} & Yes, from what I can see in the image, the chicken pieces appear to have a nice golden-brown color and look properly cooked through. You can see they've lost their raw pink color and have taken on a light golden hue. They're being removed from the pan with a black spoon, which is exactly what you should be doing at this step. The bits of browned fond left in the pan will add great flavor to the next steps of the recipe. \\[3ex]  \midrule

\multirow{2}{*}{\raisebox{-0.85\height}{\includegraphics[width=0.2\textwidth]{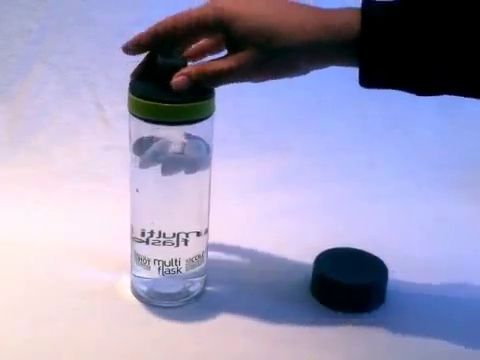}}} & \textbf{Question:} & Is the green and black cap shown in the image the correct type of cap for sealing a volumetric flask? \\[4ex]
& \textbf{GT Answer:} & Yes, that appears to be the correct type of cap for a volumetric flask. The cap shown in the image has a green band and black top which is a standard laboratory flask cap design. It's designed to create an airtight seal when properly attached to the volumetric flask. Make sure to screw it on firmly but not too tight to avoid damaging the threads. \\[1ex]
\bottomrule
\end{tabular}
\end{table*}

\section{Dialogue-level Evaluation Details}
\label{app:dialogue_level}

To address the lack of a dialogue level measurement of model quality we conducted an additional test against the strongest baselines. As most baselines do not support CVMR, we collected all dialogues from the test set that did not contain any CVMR turns, this resulted in 54 dialogues out of 340. For each dialogue, the model was tasked with answering each user request in a turn-by-turn manner, with previous responses being added to the dialogue history and model context. This allows mistakes to be reflected in the context and propagate through the dialogue, thus impacting overall dialogue performance. As user simulation with dynamic image inputs is not feasible, we opted to use the dialogue's original user requests as the user inputs, while this means that the user does not acknowledge incorrect model responses, it still allows for an insight into long term model reasoning on extended conversations.

For the evaluation protocol, we opt for an LLM-as-a-judge approach as it allows for flexible implementation and support for very long inputs (full dialog traces) with differing styles; for the judge model, we used Claude 3.5 Sonnet v2~\cite{claude}. Specifically, we tasked the model with measuring performance on a scale of 1-5 in 3 dimensions:
\begin{enumerate}
    \item \textbf{State Tracking} - This dimension seeks to measure how capable the model was at tracking user progress. This involves not only accurately understanding the user's intent with each navigation request but also possessing strong visual-text plan alignment that allows the model to infer the new state in VSG turns.
    \item \textbf{Instruction Clarity} - In a plan guidance setting it is crucial that the model is able to provide instruction in a clear manner with minimal additional and unwanted conversation filler. This dimension scores the model on how well it can keep the instructions clear and in line with user requests.
    \item \textbf{Plan Adherence} - This dimension measures how strongly the model adheres to the plan without hallucinating additional steps, ingredients, or tools.
\end{enumerate}

The complete prompt is shown in Table~\ref{tab:diag_eval_prompt}.

\begin{table*}[h]
\centering
\caption{Prompt used to evaluate the dialogues in the dialogue level evaluation described in Section~\ref{sec:dialogue_level_eval}}
\label{tab:diag_eval_prompt}
\small
{\texttt{%
\begin{tabular}{p{0.98\linewidth}}
\toprule
You are an expert evaluator for Multimodal Procedural Guidance Assistants.\\
Your task is to evaluate a full dialogue session between a User and a VLM Assistant.\\
\\
For the purpose of this evaluation you are provided with:\\
- TASK: The official ground-truth procedural plan (recipe, manual, etc.).\\
- Dialogue Transcript: A chronological log of the interaction. Each turn contains:\\
   - User Input: Text and (optional) Image Description.\\
   - Assistant Response: The response you must evaluate.\\
   - Ground-Truth Response: The ground truth response (useful for checking state/facts). DO NOT use this to score the assistant's style or tone, only for state verification.\\
\\
Your task is to evaluate the entire interaction on these 3 dimensions (Score 1-5, where 1 is poor, 3 is acceptable, and 5 is excellent):\\
\\
1 - State Tracking\\
   - Does the Assistant correctly identify the user's progress through the TASK based on the User's text and images?\\
   - If the User silently skips steps (evident in images or ground-truth response), does the Assistant correctly recognize this and jumps to the new step?\\
   - Penalize drifting from the plan or failing to recognize visual completion of steps, but do so proportionally to the severity of the error.\\
\\
2 - Instruction Clarity\\
   - Does the Assistant provide direct, actionable instructions without excessive conversational filler or unwanted commentary?\\
   - The goal is that the Assistant should be a tool, not a chatty companion. It should only speak when necessary to guide or warn, and prioritize instruction clarity.\\
   - Do not penalize politeness or empathy.\\
\\
3 - Plan Adherence\\
   - Does the Assistant remain faithful to the TASK?\\
   - Penalize hallucinating tools, ingredients, or steps that do not exist in the TASK.\\
\\
OUTPUT FORMAT\\
1 - First, provide a concise reasoning block analyzing the dialogue. You may critique specific turns.\\
2 - End with a valid JSON block, with the following structure:\\
\{\{\\
  "state\_tracking\_score": int,\\
  "succinctness\_score": int,\\
  "plan\_adherence\_score": int,\\
\}\}\\
\\
Here is the TASK:\\
\{task\_description\}\\
\\
Here is the dialogue:\\
\{dialogue\}\\
\\
Please provide your evaluation below.\\
\bottomrule
\end{tabular}}}
\end{table*}

\section{Qualitative Examples}

In Tables~\ref{tab:qual_examples_1},~\ref{tab:qual_examples_2}, and~\ref{tab:qual_examples_3} we share some examples of generated answers to plan-grounded pVQA turns from~\datasetname{} test set. In Tables~\ref{tab:qual_examples_4} and ~\ref{tab:qual_examples_5} we compare how different model sizes answer the same user questions, highlighting their capabilities and limitations.

\begin{table*}[!tbp]
\caption{Examples of answers provided by \modelname{} and some baselines to pVQA samples. In this example LLaVA-OV provided a very dangerous suggestion to the user, whereas both \modelname{} and Qwen2.5 VL were able to provide adequate and safe answers.}
\label{tab:qual_examples_1}
\centering
\begin{tabular}{l|p{0.8\textwidth}}
\toprule
\textbf{User Image} & \includegraphics[width=0.4\textwidth]{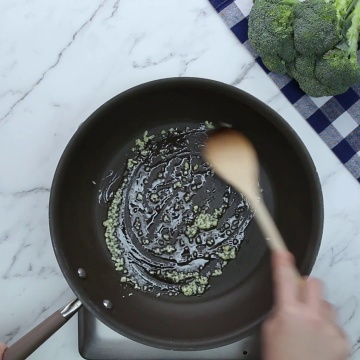} \\ \midrule
\textbf{Question:} & I've added the oil and garlic to the pan as shown in the image, but how do I know if the heat is at the right temperature? I can see some garlic starting to sizzle but I don't want to burn it. \\[4ex]\midrule 
\textbf{VIGIA 8B:} & The heat appears to be at a good medium temperature as shown in the image. You can see the garlic is starting to sizzle and release its oil, which is the right reaction. If you're seeing this gentle sizzling and the garlic isn't browning too quickly, the heat is appropriate. Just keep an eye on it and stir occasionally to prevent burning. The garlic should be fragrant but not browned when you move on to the next step. \\[4ex] \midrule 
\textbf{Qwen2.5VL:} & To ensure the heat is at the right temperature, you can use a few methods:\\
& \\
& 1. **Sizzling Garlic:** As you've noticed, the garlic is sizzling in the pan. This is a good sign because it indicates the oil is hot enough to start cooking the garlic without burning it. However, if the sizzling becomes too intense or the garlic starts to brown quickly, reduce the heat slightly.\\
& \\
& 2. **Visual Check:** If you can see the oil moving around the pan and the garlic pieces are moving as well, it’s likely the oil is at a good temperature. \\[4ex]\midrule 
\textbf{LLaVA-OV:} & To check if the heat is at the right temperature, you can use the "tongue test." Place the tip of your tongue close to the pan, but not touching it. If it feels warm but not uncomfortable, the heat is likely medium. If it feels too hot, reduce the heat. If it feels cool, increase the heat. Remember, it's better to start with a lower heat and adjust as needed to prevent burning.\\[4ex]
\bottomrule
\end{tabular}
\end{table*}

\begin{table*}[!tbp]
\caption{Examples of answers provided by \modelname{} and some baselines to pVQA samples. In this example, all three models provided adequate answers but general-purpose models tend to be more verbose.}
\label{tab:qual_examples_2}
\centering
\begin{tabular}{l|p{0.8\textwidth}}
\toprule
\textbf{User Image} & \includegraphics[width=0.4\textwidth]{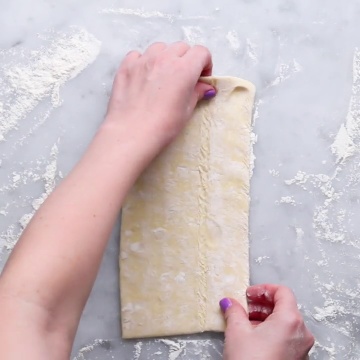} \\ \midrule
\textbf{Question:} & How do I know if I've rolled the pastry out to the correct dimensions? Should I measure it or is eyeballing it okay? \\[1ex]\midrule 
\textbf{VIGIA 8B:} & For best results, you should measure your rolled pastry to ensure it's exactly 24 inches long and 8 inches wide. This is crucial because puff pastry needs to be rolled to the correct size to properly fold and layer. While eyeballing might work for some recipes, for puff pastry, precise measurements are essential to achieve the right texture and layers. You can use a measuring tape or ruler to check the dimensions while rolling. \\[1ex] \midrule 
\textbf{Qwen2.5VL:} & When rolling out the puff pastry, it's important to ensure that it reaches the desired dimensions to achieve the best results. Here’s how you can check:\\
& \\
& \#\#\# Measuring:\\
& - **Use a Ruler:** Place a ruler next to the pastry and measure the length and width. This ensures accuracy and consistency. \\
& - **Mark the Dimensions:** If you're using a recipe, mark the dimensions on the pastry sheet with a pen or a small piece of paper to guide your rolling. \\
& \\
& \#\#\# Eyeballing: \\
& - **Experience:** If you have experience with baking and are comfortable with your eye, you might be able to eyeball it. \\[1ex]\midrule 
\textbf{LLaVA-OV:} & When rolling out puff pastry, it's important to ensure that the dimensions are correct for the recipe you're following. Here are a few ways to check if you've rolled the pastry out to the correct dimensions:\\
& \\
& 1. **Measurement**: Yes, you should measure the pastry. Use a ruler or a measuring tape to ensure that the length and width of the rolled-out pastry are as specified in the recipe. This is crucial for puff pastry, as the thickness and size can affect the final texture and rise of the pastry.\\
& \\
& 2. **Visual Inspection**: While it's not as accurate as measuring, you can also visually inspect the pastry. \\[1ex]
\bottomrule
\end{tabular}
\end{table*}

\begin{table*}[!tbp]
\caption{Examples of answers provided by \modelname{} and some baselines to pVQA samples. In this example, all three models provided adequate answers but general-purpose models tend to be more verbose.}
\label{tab:qual_examples_3}
\centering
\begin{tabular}{l|p{0.8\textwidth}}
\toprule
\textbf{User Image} & \includegraphics[width=0.5\textwidth]{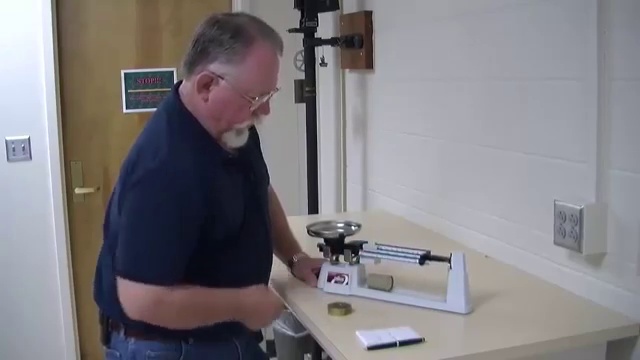} \\ \midrule 
\textbf{Question:} & I see the balance pointer is not quite centered at zero - should I adjust the calibration knob before weighing anything? \\[4ex]\midrule 
\textbf{VIGIA 8B:} & Yes, you should definitely calibrate the balance first. The image shows the balance is currently not zeroed, which means you'll need to adjust the calibration knob to ensure the indicator needle is centered at zero before proceeding to weigh any objects. This is a crucial first step to ensure accurate measurements throughout the process. \\ [4ex] \midrule 
\textbf{Qwen2.5VL:} & Yes, it's important to ensure that the balance is properly calibrated before using it for any measurements. If the pointer is not centered at zero, it means the balance is not calibrated correctly, which could lead to inaccurate readings. Here’s what you should do:\\
& \\
& 1. **Locate the Calibration Knob**: Typically, there will be a small knob or screw near the base of the balance, often labeled "Calibration" or something similar. This is used to adjust the balance.\\
& \\
& 2. **Adjust the Calibration Knob**: Turn the knob slowly in the direction needed to bring the pointer back to the zero mark. Usually, turning clockwise \\ [4ex] \midrule 
\textbf{LLaVA-OV:} & Yes, you should adjust the calibration knob to ensure the balance is properly calibrated and reading zero before weighing anything. This is a crucial step to ensure the accuracy of the weight measurement. \\[4ex]
\bottomrule
\end{tabular}
\end{table*}

\begin{table*}[!tbp]
\caption{Examples of answers provided by \modelname{} and its variants. All 3 model sizes available are capable of providing an accurate and coherent answer to the user question, while also providing a brief explanation of the importance of the thickness of the cinnamon rolls.}
\label{tab:qual_examples_4}
\centering
\begin{tabular}{l|p{0.8\textwidth}}
\toprule
\textbf{User Image} & \includegraphics[width=0.5\textwidth]{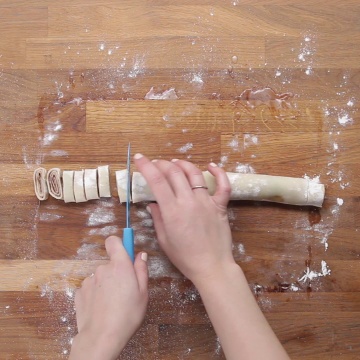} \\ \midrule 
\textbf{Question:} & Are my cinnamon roll discs being cut at the right thickness based on what's shown in the image? \\[4ex]\midrule 
\textbf{VIGIA 8B:} & Looking at the image, the cinnamon roll discs appear to be cut at a good thickness. You can see they're being cut into uniform, bite-sized pieces that are approximately 1-2 inches in diameter. This size is ideal for mini cinnamon roll bites as it ensures they'll cook evenly and be easy to eat. Just make sure to keep an eye on them while baking as the thickness may vary slightly depending on how you cut them. \\ [4ex] \midrule 
\textbf{VIGIA 3B:} & From what I can see in the image, the thickness appears to be appropriate. The discs are being cut into small, uniform pieces that are about 1-2 inches in diameter, which is ideal for mini cinnamon roll bites. This size will allow them to cook evenly and be easy to handle. When baked, these will become crispy on the outside while remaining soft on the inside. \\ [4ex] \midrule 
\textbf{VIGIA 1B:} & Yes, the thickness shown in the image appears to be appropriate for the recipe. The discs should be cut to be about 1/4 inch (6mm) thick, which is ideal for baking. If you cut them too thick, they might burn easily, and if too thin, they might burn too quickly. The uniform, flat shape shown in the image is also a good sign that you're working on the right material. \\[4ex]
\bottomrule
\end{tabular}
\end{table*}

\begin{table*}[!tbp]
\caption{Examples of answers provided by \modelname{} and its variants. The smaller model variants fail to accurately answer the user question, claiming that only a jar lid can be used whereas VIGIA 8B accurately clarifies that other objects can be used as long as they are dry and clean.}
\label{tab:qual_examples_5}
\centering
\begin{tabular}{l|p{0.8\textwidth}}
\toprule
\textbf{User Image} & \includegraphics[width=0.5\textwidth]{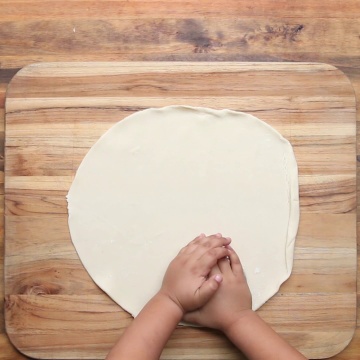} \\ \midrule 
\textbf{Question:} & The pie crust in the image looks quite round - is a jar lid the only tool I can use to cut these circles, or can I use something else of similar size? \\[4ex]\midrule 
\textbf{VIGIA 8B:} & While a jar lid is the traditional tool for cutting pie crust circles, you can use other round objects of similar size as a substitute. Some alternatives include a clean, dry glass or a small bowl. Just make sure the object you use is completely dry and clean to prevent sticking. The most important thing is to have a consistent size for your circles so they'll bake evenly. \\ [4ex] \midrule 
\textbf{VIGIA 3B:} & A jar lid is the only tool that will work for cutting the circular pie crust. The size of the jar lid (4 inches) is perfect for creating uniform, bite-sized pieces. Using a different tool could lead to unevenly sized pieces, which would be difficult to work with. Make sure you're using a clean, sharp jar lid and press it into the center of the pie crust to ensure a clean cut. \\ [4ex] \midrule 
\textbf{VIGIA 1B:} & A jar lid is the correct tool for cutting the 12 4-inch (10 cm) circles. While the image shows a circular shape, a jar lid is specifically designed for this purpose. Using a jar lid will give you the most precise and uniform cuts needed for the pie crust. Make sure to press firmly into the crust to ensure the circles are firmly attached. The circular shape will help the crust hold its shape when baked. \\[4ex]
\bottomrule
\end{tabular}
\end{table*}

\begin{table}[h]
\centering
\caption{Prompt used to generate the in-domain pVQA question-answer pairs.}
\label{tab:app_vqa_prompt}
\texttt{%
\begin{tabular}{p{\linewidth}}
\toprule
You are a helpful assistant simulating a user interacting with a visual task guidance system. You will be provided with a task, a conversation history, the current step's description, and an image related to that step.\\
\\
Based on the conversation context, the current step, and the image, generate one question-answer pair. The question should mimic a question a user might ask after uploading the provided image while seeking guidance on the current step, this is VERY important. The questions should be DIRECTLY related to the image and current task step, and should be relevant to the ongoing conversation. Make sure that these questions sound natural as if a real user had asked them. Use the conversation history, the task instructions and your knowledge to answer as accurately as possible.\\
Some examples of questions you might generate include asking if the mixture in the image is the correct consistency, if the oven temperature is correct, or if the bolt in the image needs to be tightened, etc.\\
\\
Output the question-answer pair in the following format:\\
\\
Q: <Question>\\
A: <Answer>\\
\\
---\\
Task:\\
\{task\_text\} \\ 
\\
Conversation History:\\
\{conv\_history\} \\ 
\\
---\\
\\
Current Step: \{step\_number\} - \{step\_text\} \\
Image: \\
\bottomrule 
\end{tabular}
} 
\end{table}

\begin{table}[h]
\centering
\caption{Prompt used to evaluate the in-domain pVQA responses.}
\label{tab:app_vqa_eval_prompt}
{\texttt{%
\begin{tabular}{p{0.98\linewidth}}
\toprule
You will be provided with an image and a bit of a dialogue. Your task is to tell me if the last system response accurately answers the last user question based on the image. Answer solely regarding the LAST user question and the last assistant response in the dialogue.\\
As I am parsing this automatically, please write your answer (YES or NO) immediately after the words "FINAL ANSWER:" in your response. Feel free to think and reason out loud in your response before you give the final answer.\\
\\
Task Context:\\
\\
\{task\_text\}\\
\\
Dialogue Context (up to 4 turns):\\
\{conv\_history\}\\
\\
Image: (provided in the request) \\
\bottomrule
\end{tabular}}}
\end{table}

\end{document}